\documentclass[runningheads]{llncs}

\usepackage{graphicx}
\usepackage{times}
\usepackage{epsfig}
\usepackage{graphicx}
\usepackage{amsmath}
\usepackage{amssymb}
\DeclareMathOperator*{\argmax}{argmax}
\usepackage[ruled,vlined]{algorithm2e}

\newcommand{\etal}{\textit{et al.}}

\begin{document}

\title{The Curious Case of Convex Neural Networks}


\author{%
 Sarath  Sivaprasad\thanks{Work done while at IIIT Hyderabad} \mbox{ }$^{1,2}$,\mbox{ }  Ankur Singh$^{1,3}$ \mbox{ }  Naresh Manwani$^{1}$, \mbox{ } Vineet Gandhi$^{1}$ \\
 $^{1}$KCIS, IIIT Hyderabad \mbox{ } $^{2}$TCS Research, Pune \mbox{ } $^{3}$IIT Kanpur\\
 \small{\texttt{sarath.s@research.iiit.ac.in}} \mbox{  } \small{\texttt{ankuriit@iitk.ac.in}} \mbox{  } \small{\texttt{\{naresh.manwani, vgandhi\}@iiit.ac.in}} 
}

\maketitle

\begin{abstract}
   This paper investigates a constrained formulation of neural networks where the output is a convex function of the input. We show that the convexity constraints can be enforced on both fully connected and convolutional layers, making them applicable to most architectures. The convexity constraints include restricting the weights (for all but the first layer) to be non-negative and using a non-decreasing convex activation function. Albeit simple, these constraints have profound implications on the generalization abilities of the network. We draw three valuable insights: (a) Input Output Convex Neural Networks (IOC-NNs) self regularize and {significantly reduce} the problem of overfitting; (b) Although heavily constrained,  they outperform the base multi layer perceptrons and achieve similar performance as compared to base convolutional architectures and {(c) IOC-NNs show robustness to noise in train labels}. We demonstrate the efficacy of the proposed idea using thorough experiments and ablation studies {on six commonly used image classification datasets} with three different neural network architectures.
\end{abstract}

\section{Introduction}

Deep Neural Networks use multiple layers to extract higher-level features from the raw input progressively. The ability to automatically learn features at multiple levels of abstractions makes them a powerful machine learning system that can learn complex relationships between input and output. Seminal work by Zhang~\etal~\cite{zhang2016understanding} investigates the expressive power of neural networks on finite sample sizes. They show that even when trained on completely random labeling of the true data, neural networks achieve zero training error, increasing training time and effort by only a constant factor. Such potential of brute force memorization makes it challenging to explain the generalization ability of deep neural networks. They further illustrate that the phenomena of neural network fitting on random labeling of training data is largely unaffected by explicit regularization (such as weight decay, dropout, and data augmentation). They suggest that explicit regularization may improve generalization performance but is neither necessary nor by itself sufficient for controlling generalization error. Moreover, recent works show that generalization (and test) error in neural networks reduces as we increase the number of parameters~\cite{neyshabur2017exploring,unifrom_conv}, which contradicts the traditional wisdom that overparameterization leads to overfitting. These observations have given rise to a branch of research that focuses on explaining the neural network's generalization error rather than just looking at their test performance ~\cite{towards_overparameter}.

\begin{figure*}[t]
    \centering
    \begin{tabular}{cccc}
    \hspace{-1em}
        \includegraphics[width=0.24\linewidth]{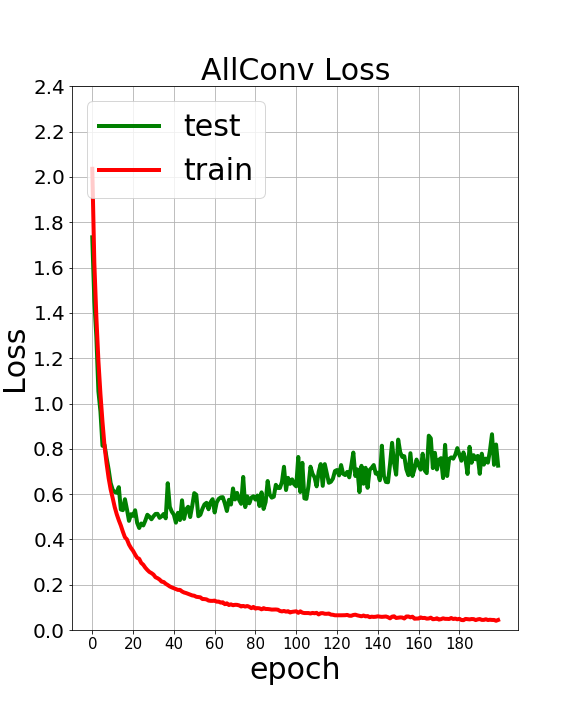}& \hspace{-1.5em} \includegraphics[width=0.24\linewidth]{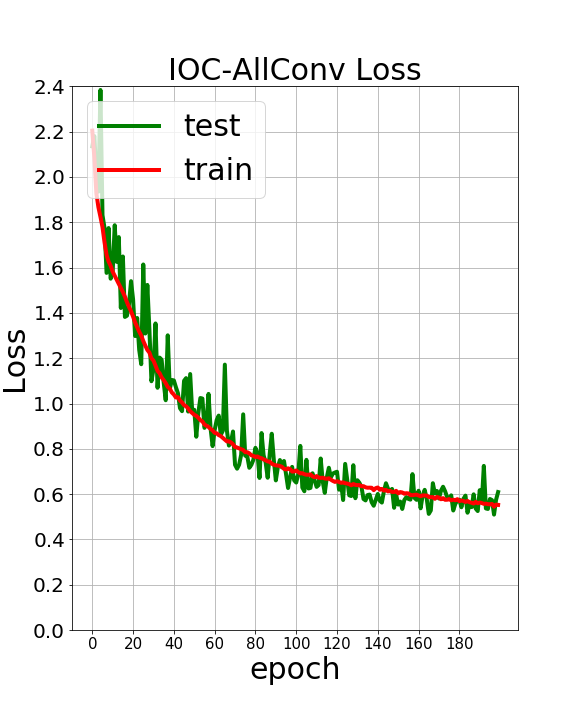}& \hspace{-1em} \includegraphics[width=0.24\linewidth]{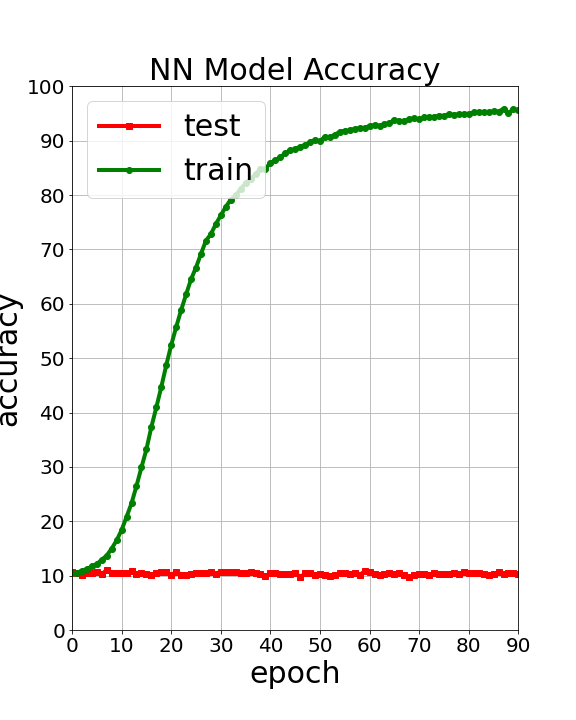}& \hspace{-1.5em} \includegraphics[width=0.24\linewidth]{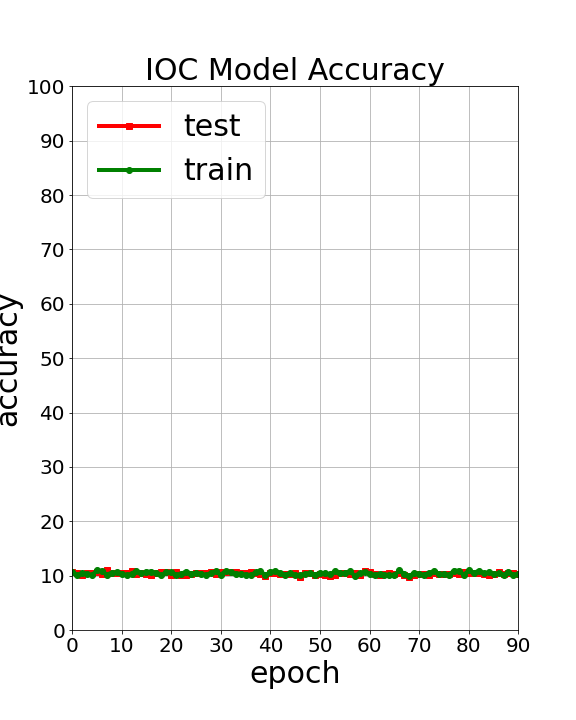} \hspace{-0.1em} \\
                \multicolumn{2}{c}{ {\small (a) True Label Experiment}} &  \multicolumn{2}{c}{ {\small(b) Random Label Experiment}} \\
        \includegraphics[width=0.24\linewidth]{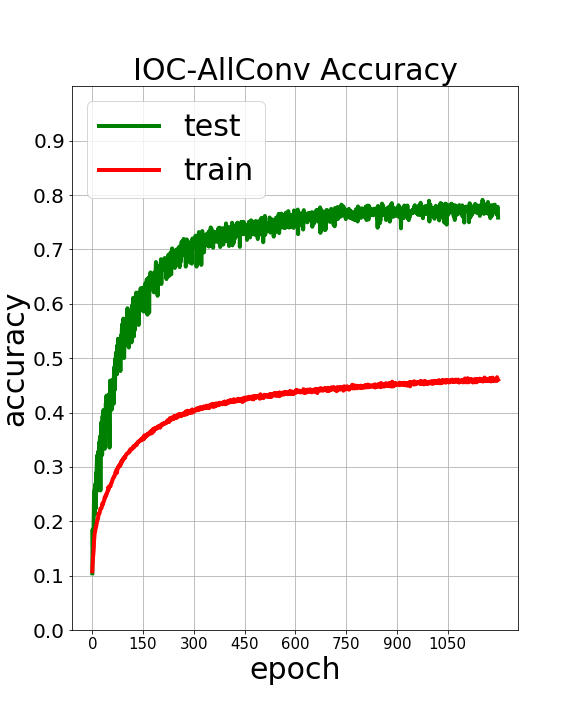}& \hspace{-1.5em} \includegraphics[width=0.24\linewidth]{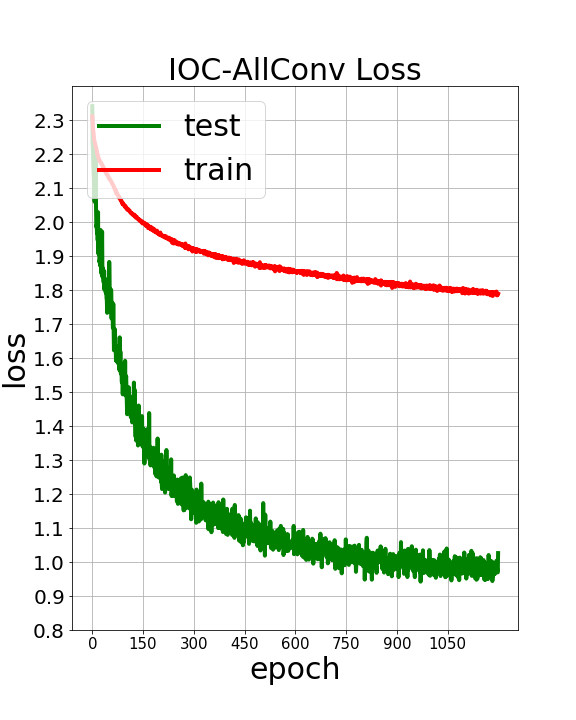}& \hspace{-1em} \includegraphics[width=0.24\linewidth]{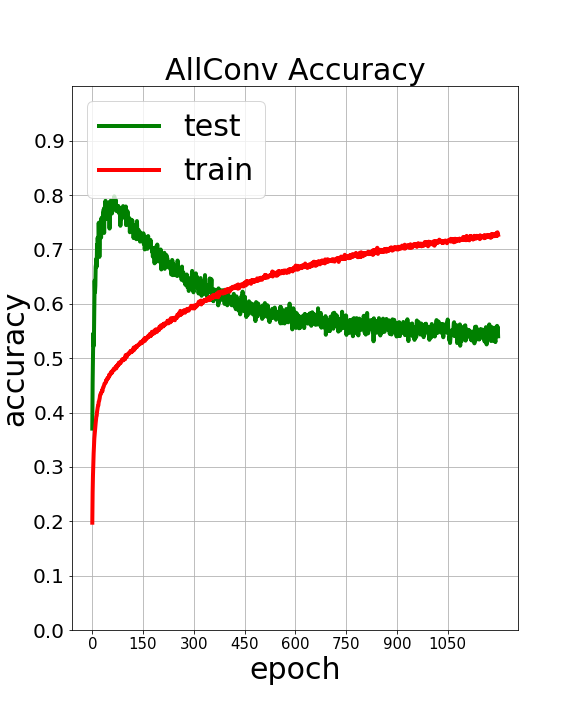}& \hspace{-1.5em} \includegraphics[width=0.24\linewidth]{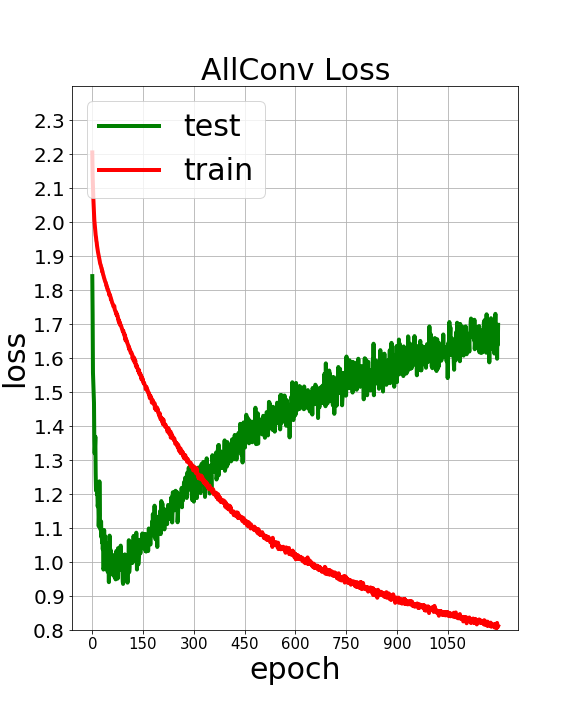} \\
          \multicolumn{2}{c}{{\small(c) IOC-AllConv (50\% noise)}} & \multicolumn{2}{c}{{\small (d) AllConv (50\% noise)}}
        \\
    \end{tabular}
    \caption{  Training of AllConv and IOC-AllConv on CIFAR-10 dataset. (a) Loss curve while training with true labels. AllConv starts overfitting after few epochs. IOC-AllConv does not exhibit overfitting, and the test loss nicely follows the training loss. (b) Accuracy plots while training with randomized labels (labels were randomized for all the training images). If sufficiently trained, even a simple network like MLP achieves 100\% training accuracy and gives around 10\% test accuracy. {IOC-MLP resists any learning on the randomized data and gives 0\% generalization gap.} (c) and (d) Loss and accuracy plots on CIFAR-10 data when trained with 50\% labels randomized in the training set.
    }
    \label{fig:teaser}
\end{figure*}

We propose a principled and reliable alternative that tries to affirmatively resolve the concerns raised in~\cite{zhang2016understanding}. More specifically, we investigate a novel constrained family of neural networks called Input Output Convex Neural Networks (IOC-NNs), which learn a convex function between input and output. Convexity in machine learning typically refers to convexity in terms of the parameters w.r.t to the loss~\cite{bach2017breaking}, which is not the case in our work. We use an IOC prefix to indicate the Input Output Convexity explicitly. Amos~\etal~\cite{amos2017input} have previously explored the idea of Input Output convexity; however, their experiments limit to Partially Input Convex Neural Networks (PICNNs), where the output is convex w.r.t some of the inputs. They deem fully convex networks \emph{unnecessary} in their studied setting of structured prediction, \emph{highly restricted} on the allowable class of models, \emph{highly limited}, even failing to do simple identity mapping without additional skip (pass-through) connections. Hence, they do not present even a single experiment on fully convex networks. 

We wake this sleeping giant up and thoroughly investigate fully convex networks (outputs are convex w.r.t to all the inputs) on the task of multi-class classification. Each class in multi-class classification is represented as a convex function, and the resulting decision boundaries are formed as an $\argmax$ of convex functions. Being able to train IOC with NN-like capacity, we, for the first time, discover the beautiful underlying properties, especially in terms of generalization abilities and robustness to label noise. We investigate IOC-NNs on six commonly used image classification benchmarks and pose them as a preferred alternative over the non-convex architectures. Our experiments suggest that IOC-NNs avoid fitting over the noisy part of the data, in contrast to the typical neural network behavior. Previous work shows that~\cite{arpit2017closer} neural networks tend to learn simpler hypotheses first. Our experiments show that IOC-NNs tend to hold on to the simpler hypothesis even in the presence of noise, without overfitting in most settings.

A motivating example is illustrated in Figure~\ref{fig:teaser}, where we train an All Convolutional network (AllConv)~\cite{springenberg2014striving} and its convex counterpart IOC-AllConv on the CIFAR-10 dataset. AllConv starts overfitting the train data after a few epochs (Figure~\ref{fig:teaser}(a)). In contrast, IOC-AllConv shows no signs of overfitting and flattens at the end (the test loss values pleasantly follow the training curve). Such an observation is consistent across all our experiments on IOC-NNs across different datasets and architectures, suggesting that IOC-NNs have {lesser reliance on explicit regularization like early stopping.} Figure~\ref{fig:teaser}(b) presents the accuracy plots for the randomized test where we train Multi-Layer Perceptron (MLP) and IOC-MLP on a copy of the data where the true labels were replaced by random labels. MLP achieves 100\% accuracy on the train set and gives a random chance performance on the test set (observations are coherent with~\cite{zhang2016understanding}). IOC-MLP resists any learning and gives random chance performance (10\% accuracy) on both train and test sets. As MLP achieves zero training error, the test error is the same as generalization error, i.e., 90\% (the performance of random guessing on CIFAR10). In contrast, the IOC-MLP has a near 0\% generalization error. We further present experiment with 50\% noisy labels Figure~\ref{fig:teaser}(c). The neural network training profile concurs with the observation of Krueger~\etal~\cite{krueger2017deep}, where the network learns a simpler hypothesis first and then starts memorizing. On the other hand, IOC-NN converges to the simpler hypothesis, showing strong resistance to fit the noise labels.

Input Output Convexity shows a promising paradigm, as any feed-forward network can be re-worked into its convex counterpart by choosing a non-decreasing (and convex) activation function and restricting its weights to be non-negative (for all but the first layer). Our experiments suggest that activation functions that allow negative outputs (like leaky ReLU or ELU) are more suited for the task as they help retain negative values flowing to subsequent layers in the network. We show that IOC-MLPs outperforms traditional MLPs in terms of test accuracy on five of the six studied datasets and IOC-NNs almost recover the performance of the base network in case of convolutional networks. In almost all studied scenarios, IOC networks achieve multi-fold improvements in terms of generalization error over unconstrained Neural Networks. Overall, our work makes the following contributions: 

\begin{itemize}
    \item We bring to light the little known idea of Input Output Convexity in neural networks. We propose a revised formulation to efficiently train IOC-NNs, retaining adequate capacity (with changes like using ELU, increasing nodes in the first layer, whitening transform at the input, etc.). To the best of our knowledge, we for the first time explore a usable form of IOC-NNs, and shows that they can be trained with NN like capacity.
    \item Through a set of intuitive experiments, we detail its internal functioning, especially in terms of its self regularization properties and decision boundaries. We show that how sufficiently complex decision boundaries can be learned using an $\argmax$ over a set of convex functions (where each class is represented by a single convex function). We further propose a framework to learn the ensemble of IOC-NNs. 
    \item With a comprehensive set of quantitative and qualitative experiments, we demonstrate IOC-NN's outstanding generalization abilities. IOC-MLPs achieve near zero generalization error in all the studied datasets and a negative generalization error (test accuracy is higher than train accuracy) in a couple of them, even at convergence. Such never seen behaviour opens up a promising avenue for more future explorations. 
    \item We explore the robustness of IOC-NNs to label noise and find that it strongly resists fitting the random labels. Even while training, IOC-NNs show no signs of fitting on noisy data and efficiently learns patterns from non noisy data. Our findings ignites explorations towards tighter generalization bounds for neural networks.
\end{itemize}

\section{Related Work}
\paragraph{Simple Convex models:} Our work relates to parameter estimation on models that are guaranteed to be convex by its construction. For regression problems, Magnani and Boyd~\cite{magnani2009convex} study the problem of fitting a convex piecewise linear function to a given set of data points. For classification problems, this traditionally translates to polyhedral classifiers. A polyhedral classifier can be described as an intersection of a finite number of hyperplanes. There have been several attempts to address the problem of learning polyhedral classifiers \cite{manwani2010learning,NIPS2014_5511}. However, these algorithms require the number of hyperplanes as an input, which is a major constraint. Furthermore, these classifiers do not give completely smooth boundaries (at the intersection of hyperplanes). As another major limitation, these classifiers cannot model the boundaries in which each class is distributed over the union of non-intersecting convex regions (e.g., XOR problem). The proposed IOC-NN (even with a single hidden layer) supersedes this direction of work. 
\paragraph{Convex Neural Networks:} Amos~\etal~\cite{amos2017input} mentions the possibility of fully convex networks, however, does not present any experiments with it. The focus of their work is to achieve structured predictions using partially convex network (using convexity w.r.t to some of the inputs). They propose a specific architecture called FICNN which is fully convex and has fully connected layers with skip connections. The skip connections are a must because their architecture cannot even achieve identity mapping without them. In contrast, our work can take any given architecture and derive its convex counterpart (we use the IOC suffix to suggest model agnostic nature of our work). The work by Kent~\etal~\cite{kent2016input} analyze the links between polynomial functions and input convex neural networks to understand the trade-offs between model expressiveness and ease of optimization. Chen~\etal~\cite{chen2018optimal,chen2020input} explore the use of input convex neural network in a variety of control applications like voltage regulation. The literature on input convex neural networks has been limited to niche tailored scenarios. Two key highlights of our work are: (a) to use activations that allow the flow of negative values (like ELU, leaky ReLU, etc.), which enables a richer representation (retaining fundamental properties like identity mapping which are not achievable using ReLU) and (b) to bring a more in-depth perspective on the functioning of convex networks and the resulting decision boundaries. Consequently, we present IOC-NNs as a preferred option over the base architectures, especially in terms of generalization abilities, using experiments on mainstream image classification benchmarks.  

\paragraph{Generalization in Deep Neural Nets:}
Conventional machine learning wisdom says that overparameterization leads to poor generalization performance owing to overfitting. Counter-intuitively, empirical evidence shows that neural networks give better generalization with an increased number of parameters even without any explicit regularization~\cite{neyshabur2015inductivebias}. Explaining how neural networks generalize despite being overparameterized is an important question in deep learning ~\cite{neyshabur2015inductivebias,unifrom_conv}.

Neyshabur ~\etal~\cite{neyshabur2017exploring} study different complexity measures and capacity bounds based on the number of parameters, VC dimension, Rademacher complexity etc., and conclude that these bounds fail to explain the generalization behavior of neural networks on overparameterization. Neyshabur~\etal~\cite{towards_overparameter} suggest that restricting the hypothesis class gives a generalization bound that decreases with an increase in the number of parameters. Their experiments show that restricting the spectral norm of the hidden layer leads to tighter generalization bounds. 

The above discussion implies that a hypothetical neural network that can fit any hypothesis will have a worse generalization than the practical neural networks which span a restricted hypothesis class. Inspired by this idea, we propose a principled way of restricting the hypothesis class of neural networks (by convexity constraints) that improves their generalization ability in practice. In the previous efforts to train fully input output convex networks, they were shown to have a limited capacity compared to its neural network counterpart~\cite{amos2017input,bach2017breaking}, making their generalization capabilities ineffective in practice. To our knowledge, we for the first time present a method to formulate and efficiently train IOC-NNs opening an avenue to explore their generalization ability. 

\section{Input Output Convex Networks}
We first consider the case of an MLP with $k$ hidden layers. The output of $i^{th}$ neuron in the $l^{th}$ hidden layer will be denoted as $h_{i}^{(l)}$. For an input $\mathbf{x}=(x_1,\ldots,x_d)$, $h_{i}^{(l)}$ is defined as:
\begin{equation}
     h_{i}^{(l)} = \phi( \sum_{j} w_{ij}^{(l)} h_j^{l-1} + b_{i}^{(l)} ),
\end{equation}
where, $h_{j}^{(0)} = x_j$ ($j=1\ldots d$) and $h_{j}^{(k+1)} = y_j$ ($j^{th}$ output). The first hidden layer represents an affine mapping of input and preserves the convexity (i.e. each neuron in $h^{(1)}$ is convex function of input). The subsequent layers are a weighted sum of neurons from the previous layer followed by an activation function. The final output $\mathbf{y}$ is convex with respect to the input $\mathbf{x}$ by ensuring two conditions: (a) $w_{ij}^{(2:k+1)} \ge 0 $ and (b) $\phi$ is convex and a non-decreasing function. The proof follows from the operator properties~\cite{boyd2004convex} that the non-negative sum of convex functions is convex and the composition $f(g(x))$ is convex if $g$ is convex and $f$ is convex and non-decreasing. 

A similar intuition follows for convolutional architectures as well, where each neuron in the next layer is a weighted sum of the previous layer. Convexity can be assured by restricting filter weights to be non-negative and using a convex and non-decreasing activation function. Filter weights in the first convolutional layer can take negative values, as they only represent an affine mapping of the input. The maxpool operation also preserves convexity since point-wise maximum of convex functions is convex~\cite{boyd2004convex}. Also, the skip connection does not violate Input Output Convexity, since the input to each layer is still a non-negative weighted sum of convex functions. 

We use an ELU activation to allow negative values; this is a minor but a key change from the previous efforts that rely on ReLU activation. For instance, with non-negativity constraints on weights ($ w_{ij}^{(2:k+1)} \ge 0 $), ReLU activations restrict the allowable use of hidden units that mirror the identity mapping. Previous works rely on passthrough/skip connections to address~\cite{amos2017input} this concern. The use of ELU enables identity mapping and allows us to use the convex counterparts of existing networks without any architectural changes. 

\subsection{Convexity as Self Regularizer}
\label{subsec:selfregularization}

We define self regularization as the property in which the network itself has some functional constraints. Inducing convexity can be viewed as a self regularization technique. For example, consider a quadratic classifier in $\mathbb{R}^2$ of the form $f(x_1,x_2)=w_1x_1^2+w_2x_2^2+w_3x_1x_2+w_4x_1+w_5x_2+w_0$. If we want the function $f$ to be convex, then it is required that the network imposes following constraints on the parameters, $w_1\geq 0,\;w_2\geq 0,\;-2\sqrt{w_1w_2}\leq w_3\leq 2\sqrt{w_1w_2}$, which essentially means that we are restricting the hypothesis space.  

Similar inferences can be drawn by taking the example of polyhedral classifiers. Polyhedral classifiers are a special class of Mixture of Experts (MoE) network \cite{10.1162/neco.1991.3.1.79,shah2019plume}. VC-dimension of a polyhedral classifier in $d$-dimension formed by the intersection of $m$ hyperplanes is upper bounded by $2(d+1)m\log(3m)$ \cite{Takacs2007}. On the other hand, VC-dimension of a standard mixture of $m$ binary experts in $d$-dimension is $O(m^4d^2)$ \cite{doi:10.1162/089976600300015367}. Thus, by imposing convexity, the VC-dimension becomes linear with the data dimension $d$ and $m\log(m)$ with the number of experts. This is a huge reduction in the overall representation capacity compared to the standard mixture of binary experts.

Furthermore, adding non-negativity constraints alone can lead to regularization. For example, the VC dimension of a sign constrained linear classifier in $\mathbb{R}^d$ reduces from $d+1$ to $d$ \cite{burges1998a,doi:10.1162/neco.2008.20.1.288}. The proposed IOC-NN uses a combination of sign constraints and restrictions on the family of activation functions for inducing convexity. The representation capacity of the resulting network reduces, and therefore, regularization comes into effect. This effectively helps in improving generalization and controlling overfitting, as clearly observed in our empirical studies (Section~\ref{section:Results}). 

\begin{figure}[t]
    \centering
    \begin{tabular}[t]{c c c c}
    \hspace{-1em}
        \includegraphics[width=0.24\linewidth]{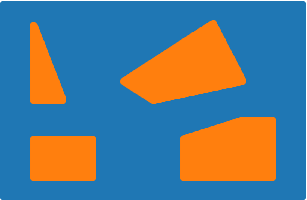}&  \includegraphics[width=0.24\linewidth]{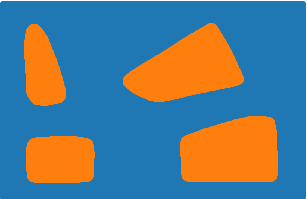}&  
        \includegraphics[width=0.24\linewidth]{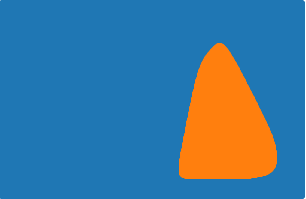}& \includegraphics[width=0.24\linewidth]{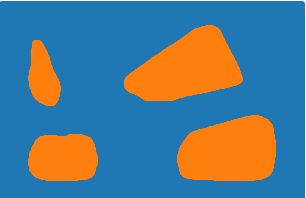} \\

        {\small (a)} &  {\small (b)} &  {\small (c)} &  {\small (d)}\\
    \end{tabular}
    \caption{ Decision boundaries of different networks trained for two class classification. (a) Original data: one class shown by blue and the other orange. (b) Decision boundary learnt using MLP. (c) Decision boundary learnt using IOC-MLP with single node in the output layer. (d) Decision boundary learnt using IOC-MLP with two nodes in the output layer (ground truth as one hot vectors)}.
    \label{fig:decision_boundaries}
\end{figure}

\begin{figure}
    \centering
    \begin{tabular}[t]{ccc}
    \includegraphics[width=0.37\linewidth]{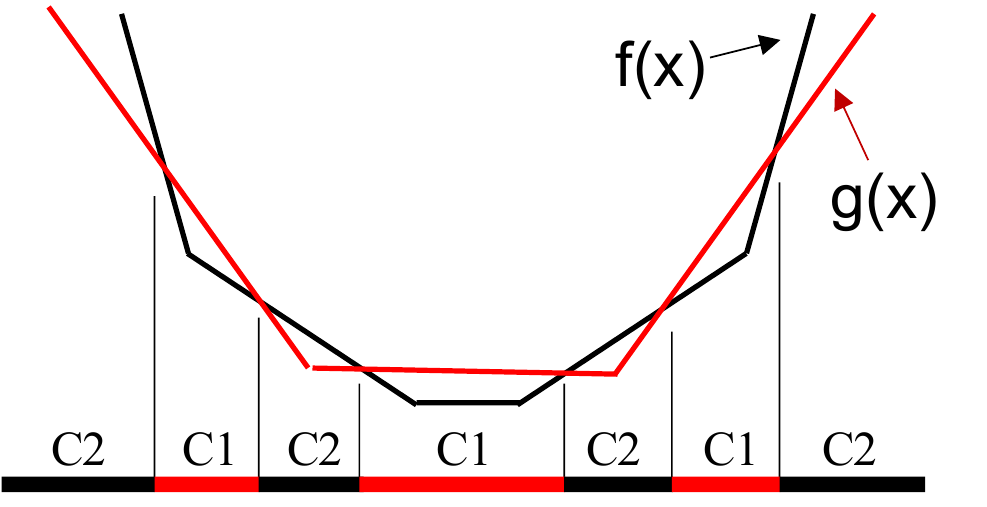}& \hspace{-0.5em} 
    \includegraphics[width=0.27\linewidth]{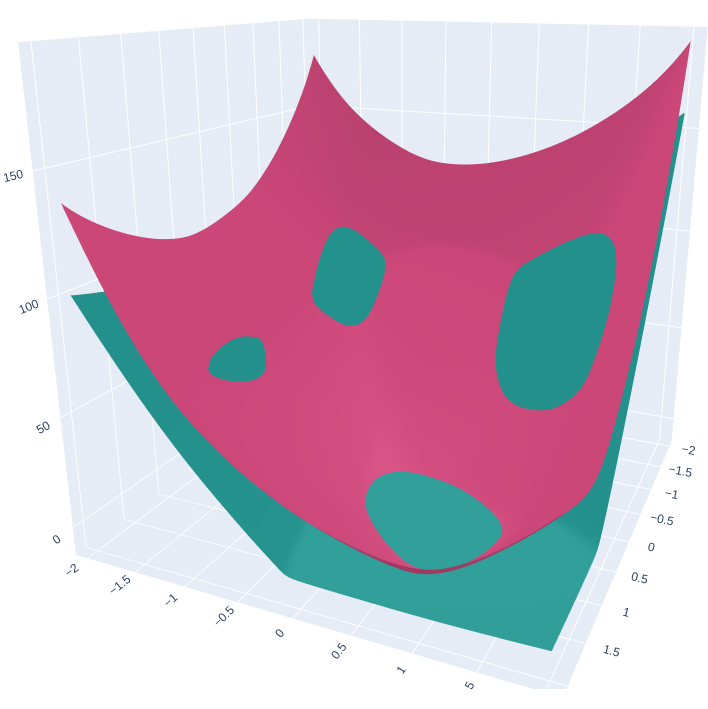} & \hspace{-1em} \includegraphics[width=0.27\linewidth]{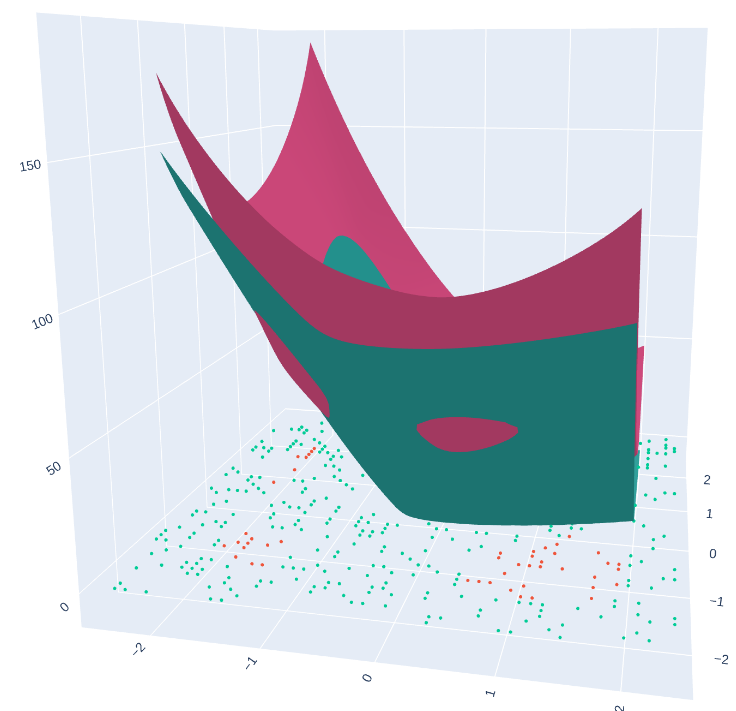} \\ \hspace{-1em}
    (a) & \multicolumn{2}{c}{(b)} \\
    \end{tabular}
    \caption{ (a) Using two simple 1-D functions we illustrate that $\argmax$ of two convex functions can result into non-convex decision boundaries. (b) Two convex functions whose $\argmax$ results into the decision boundaries shown in Figure~\ref{fig:decision_boundaries}(d). The same plot is shown from two different viewpoints. 
    }
    \label{fig:diff_convex}
\end{figure}

\subsection{IOC-NN Decision Boundaries}

Consider a scenario of binary classification in 2D space as presented in Figure~\ref{fig:decision_boundaries}(a). We train a three-layer MLP with a single output and a sigmoid activation for the last layer. The network comfortably learns to separate the two classes. The learned boundaries by the MLP are shown in Figure~\ref{fig:decision_boundaries}(b). We then train an IOC-MLP with the same architecture. The learned boundary is shown in Figure~\ref{fig:decision_boundaries}(c). IOC-MLP learns a single convex function as output w.r.t the input and its contour at the value of 0.5 define the decision boundary. The use of non-convex activation like sigmoid in the last layer does not distort convexity of decision boundary (Appendix A) 

We further explore IOC-MLP with a variant architecture where the ground truth is presented as a one-hot vector (allowing two outputs). The network learns two convex functions $f$ and $g$ representing each class, and their $\argmax$ defines the decision boundary. Thus, if $g(\mathbf{x})-f(\mathbf{x})>0$, then $\mathbf{x}$ is assigned to class $C1$ and $C2$ otherwise. Therefore, it can learn non-convex decision boundaries as shown in Figure~\ref{fig:diff_convex}. Please note that $g-f$ is no more convex unless $g''-f''\geq 0$. In the considered problem of binary classification in Figure~\ref{fig:decision_boundaries}, using one-hot output allows the network to learn non-convex boundaries (Figure~\ref{fig:decision_boundaries} (d)). The corresponding two output functions (one for each class) are illustrated in Figure~\ref{fig:diff_convex} (b). We can observe that both the individual functions are convex; however, their arrangement is such that the $\argmax$ leads to a reasonably complex decision boundary.This happens due to the fact that the sets $S_1=\{\mathbf{x}\;|\;g(\mathbf{x})-f(\mathbf{x})>0\}$ and $S_2=\{\mathbf{x}\;|\;g(\mathbf{x})-f(\mathbf{x})\leq 0\}$ can both be non-convex (even though functions $f(.)$ and $g(.)$ are convex).

\subsection{Ensemble of IOC-NN}

We further explore the ensemble of IOC-NN for multi-class classification. We explore two different ways to learn the ensembles:
\begin{enumerate}
\item Mixture of IOC-NN Experts: Training a mixture of IOC-NNs and an additional gating network~\cite{10.1162/neco.1991.3.1.79}. The gating network can be non-convex and outputs a scalar weight for each expert. The gating network and the multiple IOC-NNs (experts) are trained in an Expectation-Maximization (EM) framework, i.e., training the gating network and the experts iteratively. 
\item Boosting + Gating: In this setup, each IOC-NN is trained individually. The first model is trained on the whole data, and the consecutive models are trained with exaggerated data on the samples on which the previous model performs poorly. For bootstrapping, we use a simple re-weighting mechanism as in~\cite{friedman2000additive}. A gating network is then trained over the ensemble of IOC-NNs. The weights of the individual networks are frozen while training the gating network. 
\end{enumerate}

\begin{figure}[t]
    \centering
    \begin{tabular}[t]{ccc}
    \hspace{-1.5em}
        \includegraphics[height=0.187\linewidth]{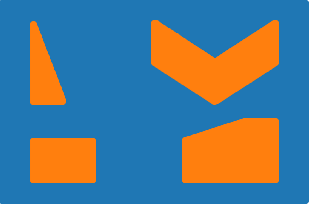}& \hspace{-1.5em} \includegraphics[height=0.19\linewidth]{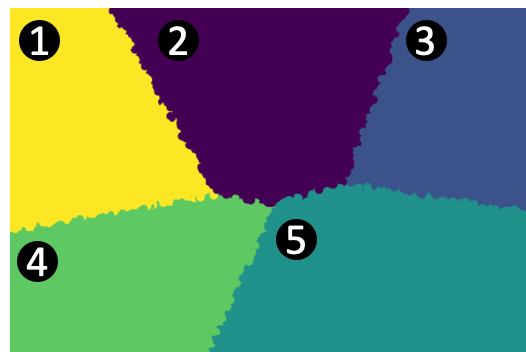}& \hspace{-1.5em}  \includegraphics[height=0.193\linewidth]{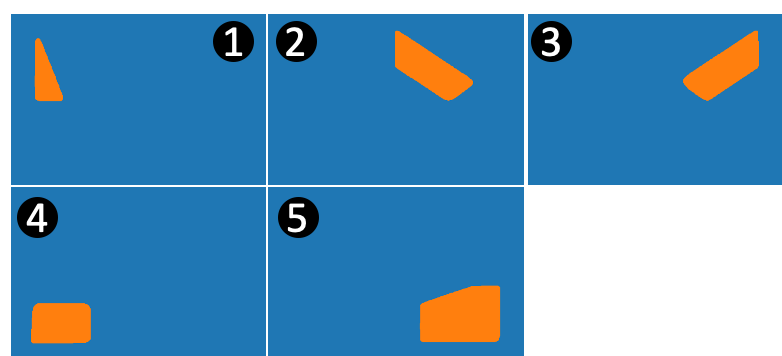} \\
        (a) & (b) & (c) \\
    \end{tabular}
    \caption{(a) Original Data. (b) Output of the gating network, each color represents picking a particular expert. (c) Decision boundaries of the individual IOC-MLPs. We mark the correspondences between each expert and the segment for which it was selected. Notice how the V-shape is partitioned and classified using two different IOC-MLPs. }
    \label{fig:ensemble}
\end{figure}

We detail the idea of ensembles using a representative experiment for binary classification on the data presented in Figure~\ref{fig:ensemble}(a). We train a mixture of $\mathbf{p}$ IOC-MLPs with a gating network using the EM algorithm. The gating network is an MLP with a single hidden layer, the output of which is a $\mathbf{p}$ dimensional vector. Each of the IOC-MLP is a three-layer MLP with a single output. We keep a single output to ensure that each IOC-MLP learns a convex decision boundary. The output of the gating network is illustrated in Figure~\ref{fig:ensemble}(b). A particular IOC-MLP was selected for each partition and led to five partitions. The decision boundaries of individual IOC-MLPs are shown in Figure~\ref{fig:ensemble}(c). It is interesting to note that the MoE of binary IOC-MLPs fractures the input space into sub-spaces where a convex boundary is sufficient for classification. 
\label{marker}

\section{Experiments}
\label{section:exp}
\paragraph{Dataset and Architectures:} 
 To show the significance of enhanced performance of IOC-MLP over traditional NN, we train them on six different datasets: MNIST, FMNIST, STL-10, SVHN,  CIFAR-10, and CIFAR-100. We use an MLP with three hidden layers and 800 nodes in each layer. We use batch normalization between every layer, and it’s activation in all hidden layers. ReLU and ELU are used as activations for NN and IOC respectively, and softmax is used in the last layer. We use Adam optimizer with an initial learning rate of 0.0001 and use validation accuracy for early stopping. 

We perform experiments that involve two additional architectures to extend the comparative study between IOC and NN on CIFAR-10 and CIFAR-100 datasets. We use a fully convolutional ~\cite{springenberg2014striving}, and a densely connected architecture~\cite{huang2017densely}. We choose DenseNet with growth rate k=12, for our experiments. We term the convex  counterparts as IOC-AllConv, IOC-DenseNet, respectively, and compare against their base neural network counterparts~\cite{huang2017densely,springenberg2014striving}. In all comparative studies, we follow the same training and augmentation strategy to train IOC-NNs, as used by the aforementioned neural networks. 

\paragraph{Training on duplicate free data:}
The test sets of CIFAR-10 and CIFAR-100 datasets have 3.25\% and 10\% duplicate images, respectively~\cite{cifair_dataset}. Neural networks show higher performance on these datasets due to the bias created by this duplicate data (neural networks have been shown to memorize the data). CIFAIR-10 and CIFAIR-100 datasets are variants of CIFAR-10 and CIFAR-100 respectively, where all the duplicate images in the test data are replaced with new images. Barz~\etal~\cite{cifair_dataset} observed that the performance of most neural architectures drops when trained and tested on bias-free CIFAIR data. We train IOC-NN and their neural network counterparts on CIFAIR-10 data with three different architectures: a fully connected network (MLP), a fully convolutional network (AllConv)~\cite{springenberg2014striving} and a densely connected network (DenseNet)~\cite{huang2017densely}.

\paragraph{Training IOC architectures:} 
We tried four variations for weight constraints to enforce convexity constraints: clipping negative weights to zero, taking absolute of weights, exponentiation of negative weights and shifting the weights after each iteration. We use exponentiation strategy in all experiments, as it gave the best results. We exponentiate the negative weights after every update. The IOC constrained optimization algorithm differs only by a single step from the traditional algorithms (Appendix B). 

To conserve convexity in the batch-normalization layer, we also constrain the gamma scaler  with exponentiation. However, in practice we found that the IOC networks retains all desirable properties without constraining the gamma scalar. We make few additional modifications to facilitate the training of IOC-NNs. Such changes do not affect the performance of the base neural networks. We use ELU as an activation function instead of ReLU in IOC-NNs. We apply the whitening transformation to the input so that it is zero-centered, decorrelated, and spans over positive and negative values equally. We also increase the number of nodes in the first layer (the only layer where parameters can take negative values). We use a slower schedule for learning rate decay than the base counterparts. The IOC-NNs have a softmax layer at the last layer and are trained with cross-entropy loss (same as neural networks).
 
\paragraph{Training ensembles of binary experts:} We divide CIFAR-10 dataset into 2 classes, namely: `Animal' (CIFAR-10 labels:  `Bird', `Cat', `Deer', `Dog', `Frog' and `Horse') and `Not Animal'. We train an ensemble of IOC-MLP, where each expert is a three-layer MLP with one output (with sigmoid activation at the output node). The gating network in the EM approach is a one layer MLP which takes an image as input and predicts the weights by which the individual expert predictions get averaged. We report test results of ensembles with each additional expert. This experiment resembles the study shown in Figure \ref{fig:ensemble}. 

\paragraph{Training Boosted ensembles:} 
The lower training accuracy of IOC-NNs makes them suitable for boosting (while the training accuracy saturates in non-convex counterparts). For bootstrapping, we use a simple re-weighting mechanism as in~\cite{friedman2000additive}. We train three experts for each experiment. The gating network is a regular neural network, which is a shallow version of the actual experts. We train an MLP with only one hidden layer, a four-layer fully convolutional network, and a DenseNet with two dense-blocks as the gate for the three respective architectures. We report the accuracy of the ensemble trained in this fashion as well as the accuracy if we would have used an oracle instead of the gating network.

\paragraph{Partially randomized labeling:} 
Here, we investigate IOC-NN's behavior in the presence of partial label noise. We do a comparative study between IOC and neural networks using All-Conv architecture, similar to the experiment performed by ~\cite{zhang2016understanding}. We use CIFAR-10 dataset and make them noisy by systematically randomizing the labels of a selected percentage of training data. We report the performance of All-Conv, and it’s IOC counterpart on 20, 40, 60, 80 and 100 percent noise in the train data. We report train and test scores at peak performance (performance if we had used early stopping) and at convergence (if loss goes below 0.001 or at 2000 epochs).

\subsection{Results}
\label{section:Results}
\paragraph{IOC as a preferred alternative for Multi-Layer-Perceptrons:}
MLP is most basic and earliest explored form of neural networks. We compare the train and test scores of MLP and IOC-MLP in Table~\ref{table:mlp_result_1}. With a sufficient number of parameters, MLP (a basic NN architecture) perfectly fits the training data. However, it fails to generalize well on the test data owing to brute force memorization. The results in Table~\ref{table:mlp_result_1} indicate that IOC-MLP gives a smaller generalization gap (the difference between train and test accuracies) compared to MLP. The generalization gap even goes to negative values on three of the datasets. MLP (being poorly optimized for parameter utilization) is one of the architectures prone to overfitting the most, and IOC constraints help retain test performance resisting the tendency to overfit. Obtaining negative or almost zero generalization error even at convergence is a never seen behaviour in deep networks and the results clearly suggest the profound generalization abilities of Input Output Convexity, especially when applied to fully connected networks.

\begin{table}[t]
\centering
\small\addtolength{\tabcolsep}{2.5pt}
\hspace*{2.6mm}\begin{tabular}{|c|c|c|c|c|c|c|}
\hline
\multicolumn{1}{|c|}{} & \multicolumn{3}{c|}{NN} & \multicolumn{3}{c|}{IOC-NN} \\ \cline{2-7} 
\multicolumn{1}{|c|}{}   & train      & test    & gen. gap      & train        & test   &gen. gap        \\ \hline
MNIST  &  99.34 & 99.16 & 0.19 &98.77 &\bf{99.25} &\bf{-0.48}
 \\ \cline{2-7}   \hline
FMNIST      & 
94.8 &\bf{90.61} &3.81 &90.41 &90.58 &\bf{-0.02}
 \\ \cline{2-7}   \hline
STL-10      & 
81 &52.32 &28.68 &62.3 &\bf{54.55} &\bf{7.75}
 \\ \cline{2-7}   \hline
SVHN      & 
91.76 &86.19 &5.57 &81.18 &\bf{86.37} &\bf{-5.19}
 \\ \cline{2-7}   \hline
CIFAR-10    & 
97.99 &63.83 &34.16 & 73.27 &\bf{69.89} &\bf{3.38}
 \\ \cline{2-7} 
\hline
CIFAR-100 & 84.6 &32.68 &51.92& 46.9 &\bf{41.08} &\bf{5.82}
\\ \cline{2-7} 
\hline
\end{tabular}
\caption{Table shows train accuracy, test accuracy and generalization gap for MLP and IOC-MLP on six different datasets.}
\label{table:mlp_result_1}
\end{table}

\begin{table*}[t]
    \hspace{-5mm}
    \centering
    \begin{tabular}{|c|c|c|c|c|c|c|c|c|c|c|c|c|}
    \hline
    &\multicolumn{6}{|c|}{CIFAR-10} & \multicolumn{6}{|c|}{CIFAR-100}\\
    \hline
    & \multicolumn{3}{|c|}{NN} & \multicolumn{3}{c|}{IOC-NN} & \multicolumn{3}{|c|}{NN} & \multicolumn{3}{c|}{IOC-NN} \\ 
    \hline
    &train      & test       & gen. gap        & train      & test       & gen. gap  & train      & test       & gen. gap        & train      & test & gen. gap       \\ \hline
    
    MLP & 99.17 & 63.83 & 35.34 & 73.27 & 69.89 & \textbf{3.3} & 84.6 &32.68 &51.9 &46.9 &41.08 &\textbf{5.8}
     \\
    \hline
    AllConv & 99.31 & 92.8 & 6.5 & 93.2 & 90.6 & \textbf{2.6} & 97.87 &69.5 & 28.4 &67.07 &65.08 &\textbf{1.9}
     \\
    \hline
    DenseNet & 99.46 & 94.06 & 5.4 & 94.22 & 91.12 & \textbf{3.1} & 98.42 &75.36 &23.06 &74.9 &68.53 &\textbf{6.3}
     \\
    \hline
    \end{tabular}
    \caption{Train accuracy, test accuracy and generalization gap of three neural architectures and their IOC counterparts}
    \label{table:main_result}
\end{table*} 

\begin{table*}[t]
\centering
\hspace*{1mm}\begin{tabular}{|c|c|c|c|c|c|c|c|c|c|c|}
\hline
&\multicolumn{5}{|c|}{NN} & \multicolumn{5}{|c|}{IOC-NN}\\
\hline
& \multicolumn{2}{|c|}{peak} & \multicolumn{3}{c|}{convergence} & \multicolumn{2}{|c|}{peak} & \multicolumn{3}{c|}{convergence} \\ 
\hline
&train      & test      &train      & test  & gen. gap        & train      & test      & train      & test  & gen. gap  \\ \hline
100&
98.63 & 
10.53 &
97.80 &
10.1&
\bf{87.7} &
9.98 &
10.62 &
10.21 &
9.94 &
\bf{0.27}
\\
\hline
80 & 
22.40 &
60.24 &
97.83&
27.75 &
\bf{70.08}&
21.93 &
61.48 &
23.80 &
56.20 &
\bf{-32.4}
\\
\hline
60 & 
38.52 &
75.80 &
97.80&
46.71 &
\bf{51.09}&
37.90 &
75.91 &
39.31 &
71.75 &
\bf{-32.44}
\\
\hline
40 & 
56.48 &
80.47 &
97.96&
61.83 &
\bf{36.13}&
55.01 &
81.58 &
54.63 &
81.01 &
\bf{-26.38}
\\
\hline
20 & 
72.8 &
85.72 &
98.73&
76.31 &
\bf{22.42}&
69.92 &
85.85 &
70.22 &
83.61 &
\bf{-13.39}
\\
\hline
\end{tabular}
\caption{Results for systematically randomized labels at peak and at convergence for both IOC-NN and NN. The IOC constraints bring huge improvements in generalization error and test accuracy at convergence.}
\label{table:noise}
\end{table*}

\begin{table*}[t]
\begin{center}
\begin{tabular}{| c | c | c | c | c | c |}
\hline
 
& \hspace{0.1cm} Base MLP \hspace{0.1cm} & \hspace{0.1cm} Constrained MLP \hspace{0.1cm} & \hspace{0.1cm}  FICNN \hspace{0.1cm} & \hspace{0.1cm} IOC-NN \hspace{0.1cm}\\
\hline
train & 99.17 & 46.81 & 62.8 & 73.27  \\ \hline
test & 63.83 & 27.36 & 53.07 & 69.89 \\ \hline
gen-gap & 35.34 & 19.45 & 9.73 & 3.38  \\ \hline

\end{tabular}
\end{center}
\caption{Results comparing FICNN~\cite{amos2017input} with IOC-NN on CIFAR-10 using MLP architecture. First column shows base MLP results. Second column presents results with a convex MLP using ReLU activation. Third and final columns show the accuracies of FICNN and IOC-NN, respectively.}
\label{table_ioc_v_ficnn}
\end{table*}

\begin{figure}[t]
    \centering
    \begin{tabular}[t]{ccc}
    \hspace{-1.5em}
    \includegraphics[width=0.35\linewidth]{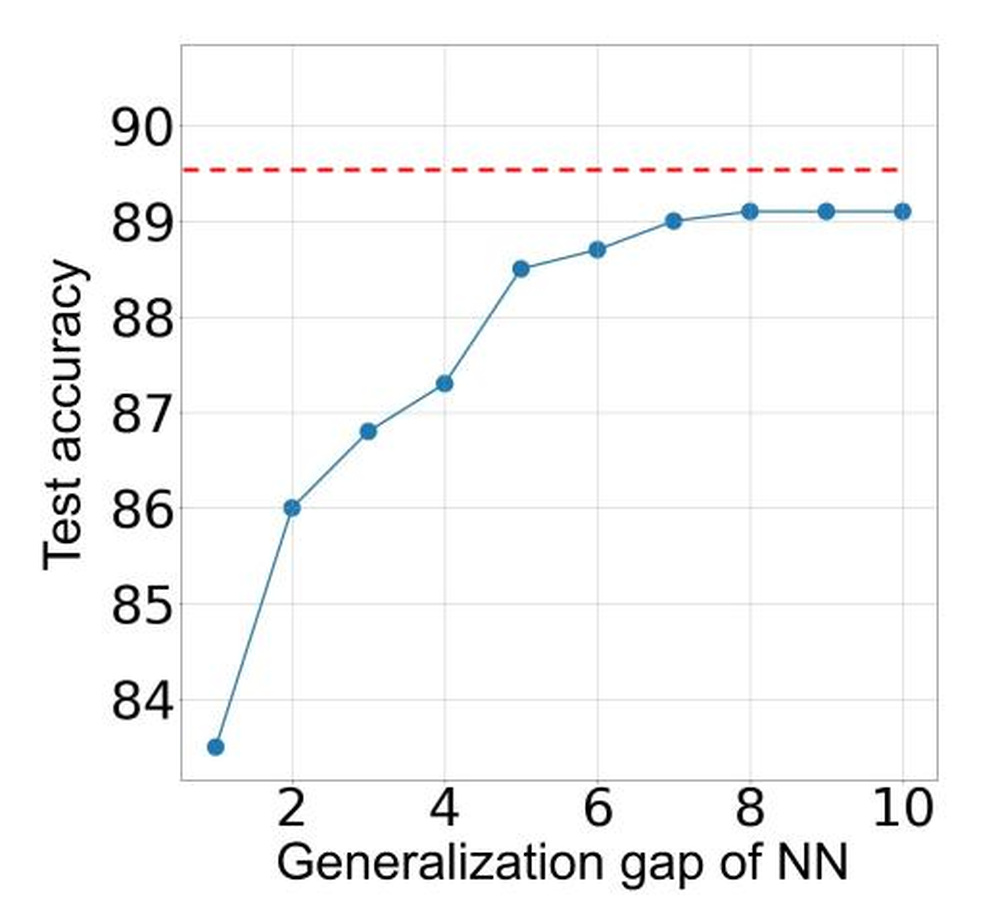}& \hspace{-0.5em} 
    \includegraphics[width=0.35\linewidth]{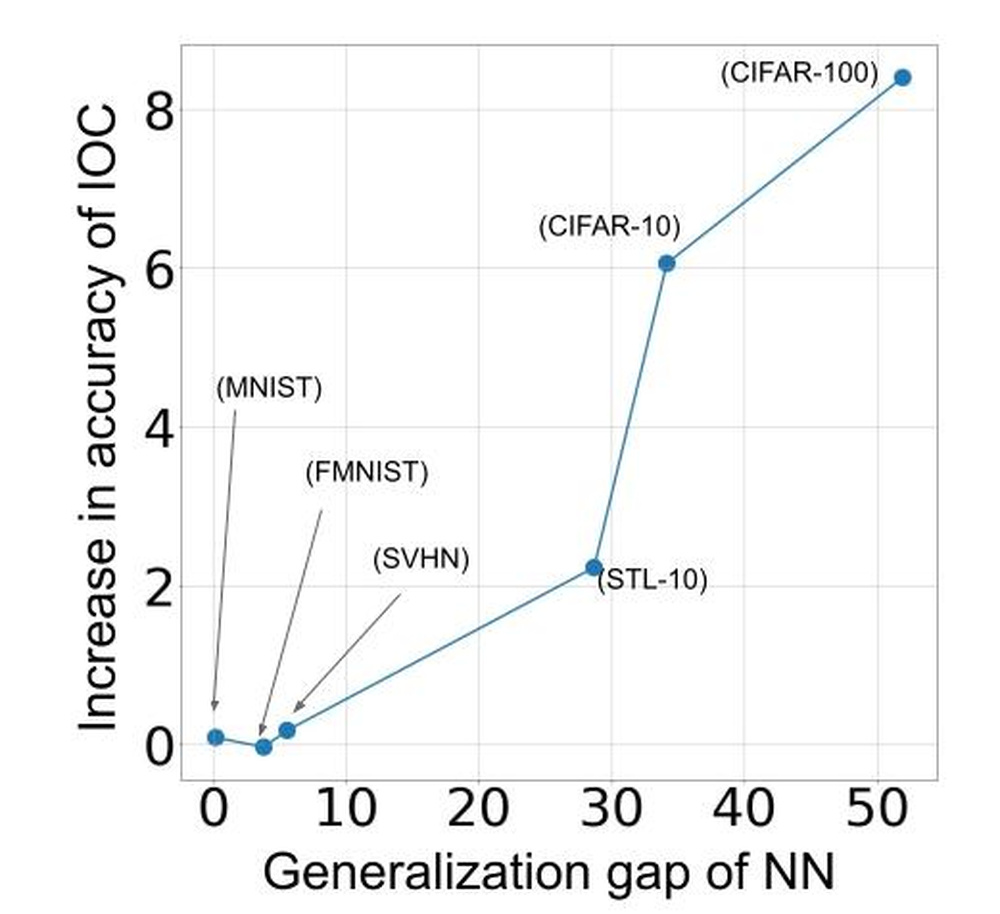}&\\
    (a) & (b) \\
    \hspace{-0.5em} \end{tabular}
    \caption{(a) shows the test accuracy of IOC-MLP with increasing number of experts in the binary classification setting. Average performance of normal MLP is shown in red since it does not change with increase in number of experts. (b) The generalization gap of MLP plotted against the improvement gained by the IOC-MLP for the six different datasets (represented by every point on the plot). The performance gain with IOC constraints increase with the increase in generalization gap of MLP.}
    \label{fig:results}
\end{figure}

Furthermore, having the IOC constraints significantly boosts the test accuracy on datasets where neural network gives a high generalization gap (Table~\ref{table:mlp_result_1}). This trend is clearly visible in Figure~\ref{fig:results}(b). For the CIFAR-10 dataset, unconstrained MLP gives 34.16\% generalization gap, while IOC-NN brings down the generalization gap by more than ten folds and boosts the test performance by about 6\%. Even in scenarios where neural networks give a smaller generalization gap (like MNIST and SVHN), IOC-NN marginally outperforms regular NN and gives an advantage in generalization. Overall, the results in Table~\ref{table:mlp_result_1} highlight that IOC constraints are extremely beneficial when training Multi Layer Perceptrons for image classification, giving comprehensive advantages in terms of generalization and test performance.

\paragraph{Better generalization:}
We investigate the generalization capability of IOC-NN on other architectures. The results of the base architectures and their convex counterparts on CIFAR-10 and CIFAR-100 datasets are presented in Table~\ref{table:main_result}. IOC-NN outperforms base NN on MLP architecture and gives comparable test accuracies for convolutional architectures. The train accuracies are saturated in the base networks (reaching above 99\% in most experiments). The lower train accuracy in IOC-NNs suggests that there might still be room for improvement, possibly through better design choices tailored for IOC-NNs. In Table~\ref{table:main_result}, the difference in train and test accuracy across all the architectures (generalization gap) demonstrates the better generalization ability of IOC-NNs. The generalization gap of base architectures is at least twofold more than IOC-NNs on the CIFAR-100 dataset. For instance, the generalization error of IOC-AllConv on CIFAR-100 is only 1.99\%, in contrast to 28.4\% in AllConv. The generalization ability of IOC-NNs is further qualitatively reflected using the training and validation loss profiles (e.g., Figure\ref{fig:teaser}(a)). We present a table showing the confidence intervals of prediction across all three architectures with repeated runs in Appendix C. 

Table~\ref{table:cifair} shows the train and test performance of the three architectures on CIFAR-10 dataset and the drop incurred when trained on CIFAIR-10. The drop in test performance of IOC-NNs is smaller than the typical neural network. This further strengthens the claim that IOC-NNs are not memorizing the training data but learning a generic hypothesis.

\paragraph{Comparison with FICNN:}
Table~\ref{table_ioc_v_ficnn} shows the results of IOC-NN and FICNN~\cite{amos2017input} on CIFAR-10 data. For comparison, we use a three layer MLP with 800 nodes in each layer, for both IOC-NN and FICNN. FICNN uses a skip connection from input layer to each of the intermittent layers. This enables each layer to learn identity mapping inspite of non-negative constraint. The number of parameters in FICNN model is almost twice compared to the base MLP and IOC models but still the test performance drops by more than  10\%. The results clearly shows that IOC-NN gives better test accuracy and lower generalization gap compared to FICNN, while using the same number of parameters as the base MLP architecture. 


\paragraph{Robustness to random label noise:}
Robustness of IOC-NNs on partial and fully randomized labels (Figure~\ref{fig:teaser} (b, c, and d)) is one of its key properties. We further investigate this property by systematically randomizing increasing portion of labels. We report the results of neural networks and their convex counterparts with percentage of label noise varying from 20\% to 100\% in Table~\ref{table:noise}. The train performance of neural networks at convergence is near 100\% across all noise levels. It is interesting to note that IOC-NN gives a large negative generalization gap, where the train accuracy is almost equal to the percentage of true labels in the data. This observation shows that IOC-NNs significantly resist learning noise in labels as compared to neural networks. Both neural network and it's convex counterpart learns the simple hypothesis first. While IOC-NN holds on to this, in later epochs, the neural network starts brute force memorization of noisy labels. The observations are coherent with findings in~\cite{krueger2017deep,sjoberg1995overtraining}, demonstrating neural network's heavy reliance on early stopping. IOC-AllConv outperforms test accuracy of AllConv + early stopping with a much-coveted generalization behavior. It is clear from this experiment that IOC-NN performs better in the presence of random label noise in the data in terms of test accuracy both at peak and convergence. 

\begin{table}[t]
\centering
\hspace*{-5mm}\begin{tabular}{| c | c | c | c | c | c | c |}
\hline
\multicolumn{1}{|c|}{}  & \multicolumn{3}{c|}{NN} & \multicolumn{3}{c|}{IOC-NN} \\ \cline{2-7} 
\multicolumn{1}{|c|}{}   & C-10      & CIFAIR    & Gap      & C-10      & CIFAIR    & Gap        \\ \hline
MLP      & 
63.6 &
63.08 &
0.52 &
69.89 &
69.51 &
\bf{0.38}
 \\ \cline{2-7}   \hline
AllConv    & 
92.8 &
91.14 &
0.66 &
90.6 &
90.47 &
\bf{0.13}
 \\ \cline{2-7} 
\hline
DenseNet & 
94.06 &
93.28 &
0.78 &
91.12 &
90.73 &
\bf{0.39}
     \\ \cline{2-7} 
\hline
\end{tabular}
\caption{Results on CIFAIR-10 dataset }
\label{table:cifair}
\end{table}

\begin{table}[t]
\centering
\hspace*{-0.8mm}\begin{tabular}{|c|c|c|c|}
\hline
& single expert & gate  & oracle \\ \hline
MLP     & 
69.89 &
71.8 &
85.47
  \\ \hline
All-Conv  & 
90.6 &
92.83 &
96.3
     \\ \hline
DenseNet   & 
91.12 &
93.25 &
97.19
 \\ \hline
\end{tabular}
\caption{Result for single expert, gated MoE and with oracle on CIFAR-10 for three architectures}
\label{table:ensemble}
\end{table}

\paragraph{Leverage IOC properties to train ensembles:}
We train binary MoE on the modified two-class setting of CIFAR-10 as described in Section~\ref{section:exp}. The result is shown in Figure~\ref{fig:results} (a). Traditional neural network gives a test accuracy of 89.63\%  with a generalization gap of 10\%. Gated MoE of NNs does not improve the test performance as we increase the number of experts. In contrast, the performance of ensemble of IOC-NNs goes up with the addition of each expert and moves closer to the performance of neural networks. It is interesting to note that even in the higher dimensional space (like CIFAR-10 images), the intuitions derived from Figure \ref{fig:ensemble} holds. We also note that gate fractures the space into \textbf{p} partitions (where p is the number of experts). Moreover, in the binary case for a single expert, the generalization gap is almost zero. This can be attributed to the convex hull like smooth decision boundary that the network predicts in the binary setting with a single output. 

The results with the boosted ensembles of IOC-NNs are presented in Table~\ref{table:ensemble}. The boosted ensemble improves the test accuracies of IOC-NNs, matching or outperforming the base architectures. However, this performance gain comes at the cost of increased generalization error (still lower than the base architectures). In the boosted ensemble, the performance significantly improves if the gating network is replaced by an oracle. This observation suggests that there is a scope of improvement in model selection ability, possibly by using a better gating architecture.

\paragraph{Confidence Calibration of IOC-NNs:}
In a classification setting, given an input, the neural network predicts probability-like scores towards each class. The class with the maximum score is considered the predicted output, and the corresponding score to be the confidence. The confidence and accuracy being correlated is a desirable property, especially in high-risk applications like self-driving cars, medical diagnoses, etc. However, many modern multi-class classification networks are poorly calibrated, i.e., the probability values that they associate with the class labels they predict overestimate the likelihoods of those class labels being correct in the real world~\cite{calibration}. Recent works have explored methods to improve the calibration of neural networks ~\cite{calibration,focal_loss}.

\begin{figure*}[t]
    \centering
    \begin{tabular}[t]{cccccc}
    \hspace{-1em}
        \includegraphics[width=0.16\linewidth]{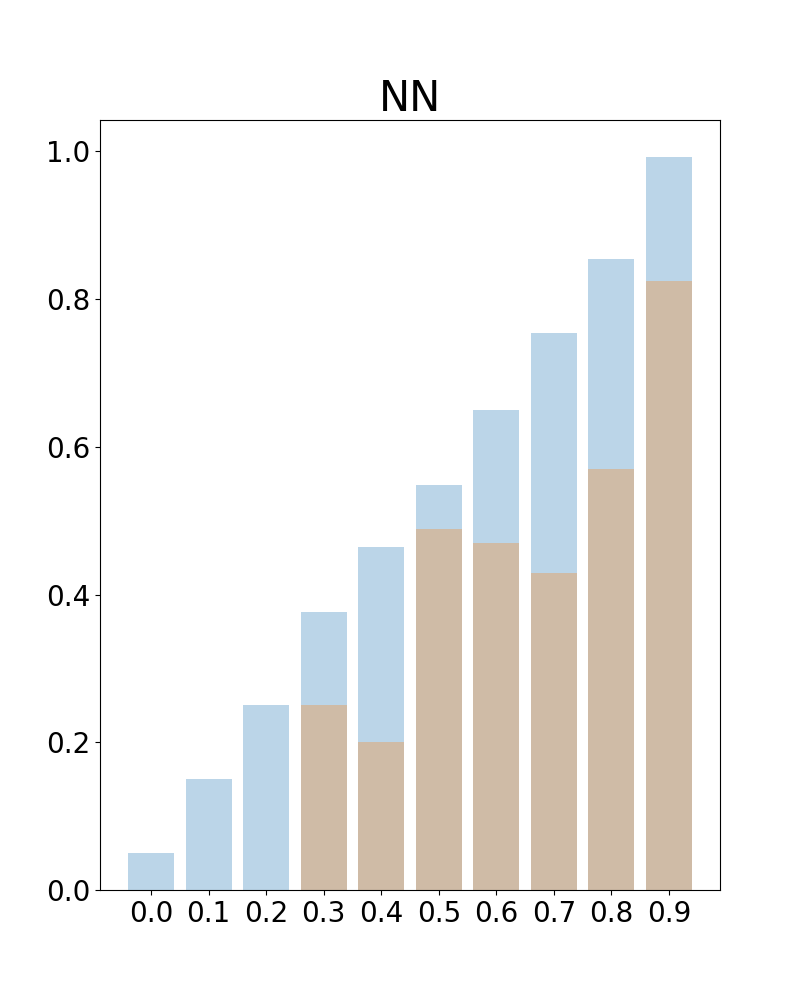}& \hspace{-0.5em} \includegraphics[width=0.16\linewidth]{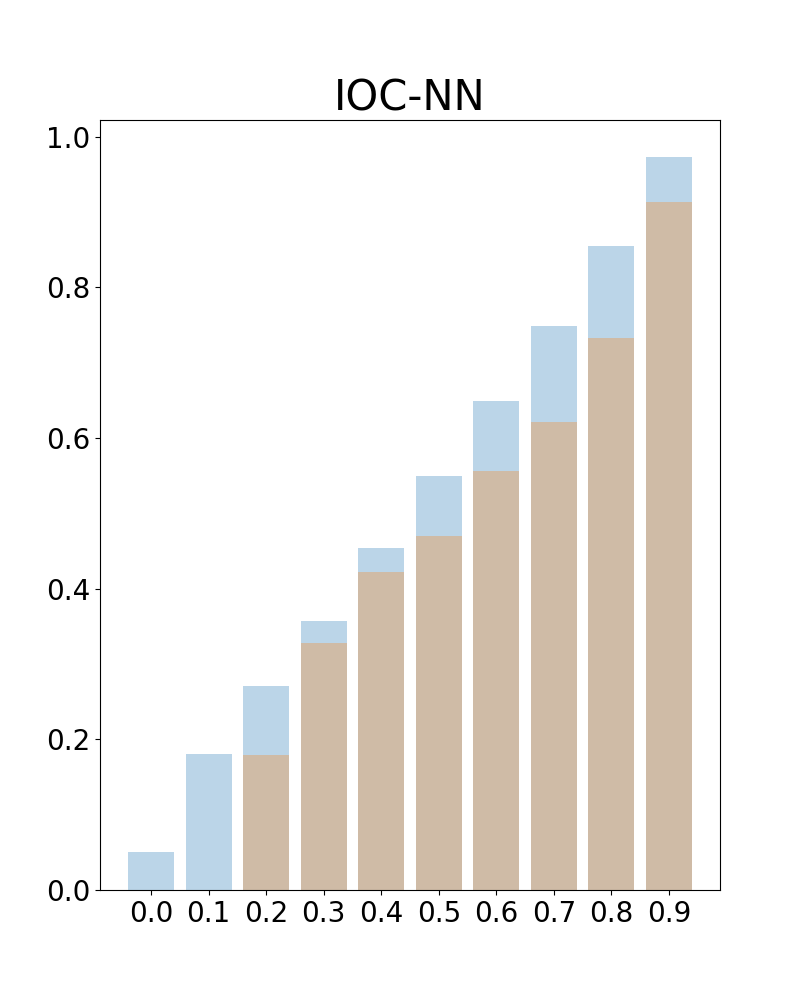}& \hspace{-0.5em} \includegraphics[width=0.16\linewidth]{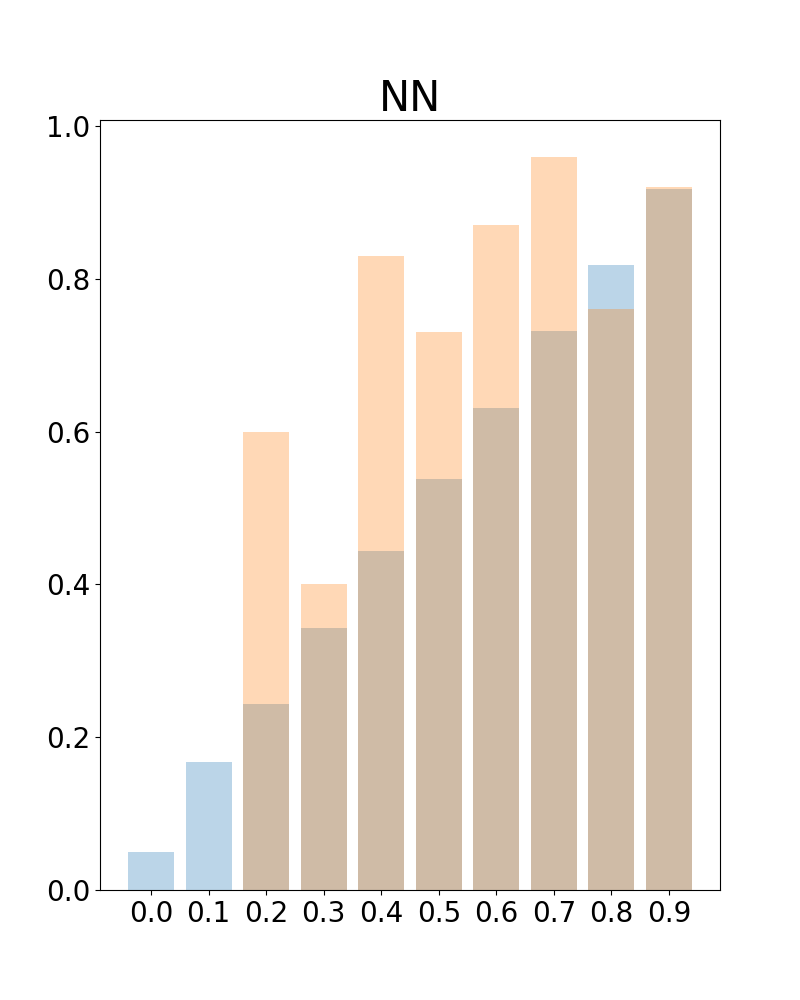}& \hspace{-0.5em} \includegraphics[width=0.16\linewidth]{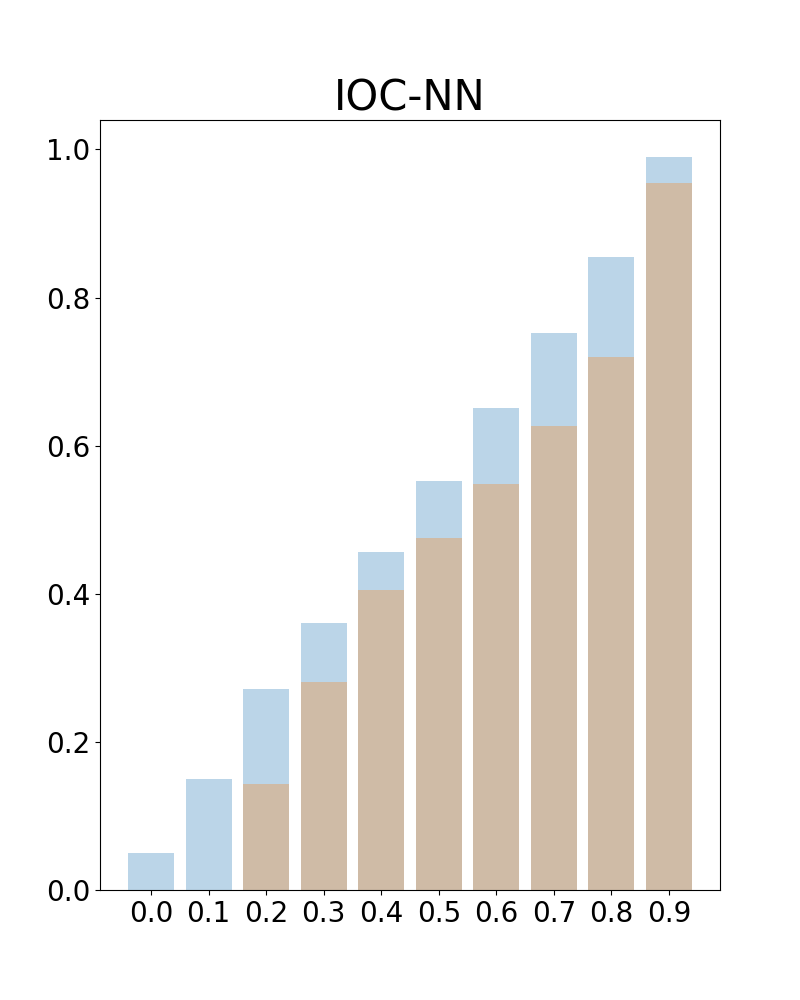} \hspace{-0.5em}
        \includegraphics[width=0.16\linewidth]{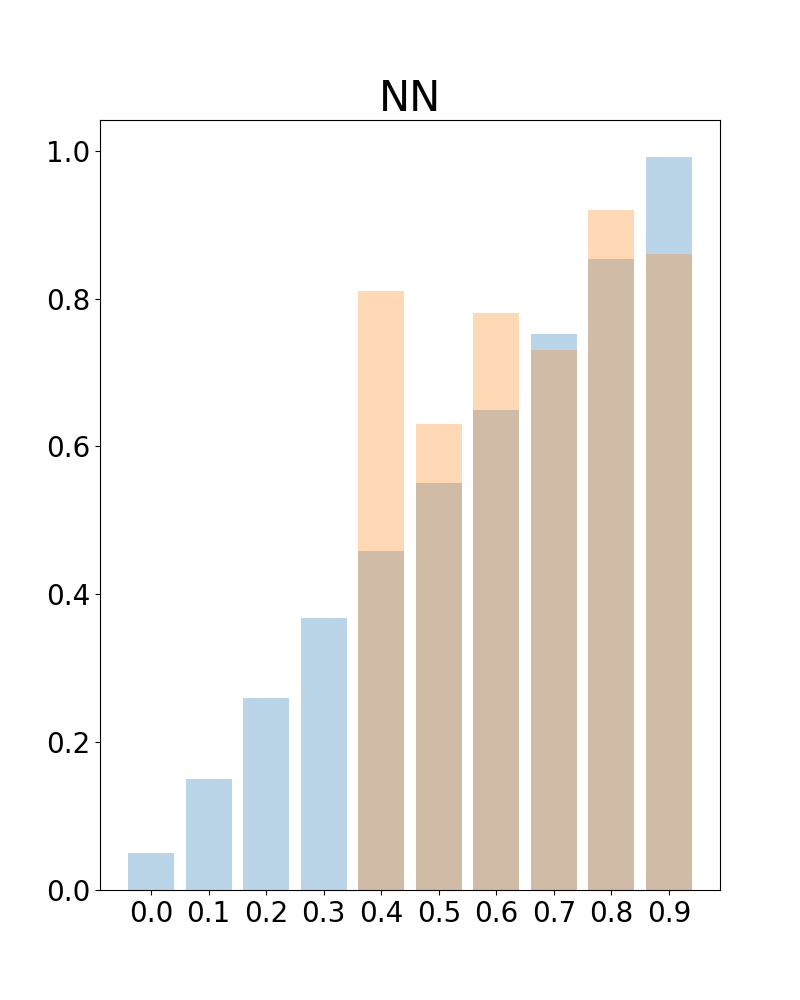}& \hspace{-0.5em} \includegraphics[width=0.16\linewidth]{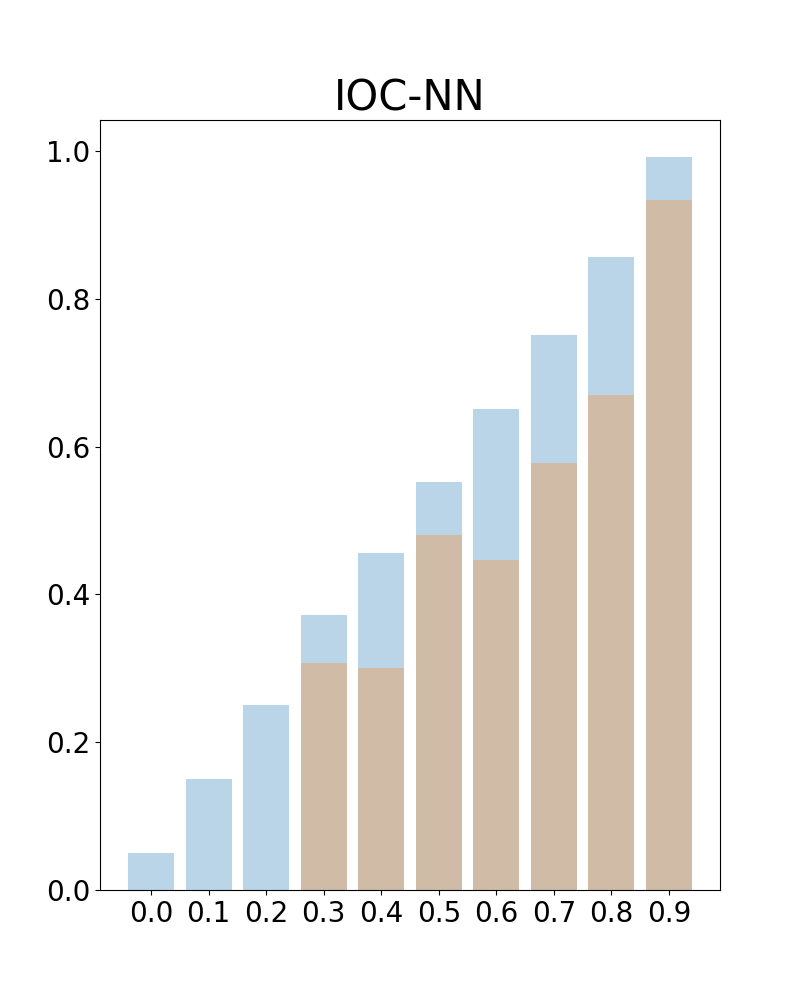}& \hspace{-0.5em} \\
         
        \multicolumn{2}{c}{\hspace{-2em} {\small (a) MLP}} &  \multicolumn{2}{c}{\hspace{-9em} {\small(b) AllConv}} &
        \multicolumn{2}{c}{\hspace{-11em} {\small(c) DenseNet}}\\
    \end{tabular}
    \caption{ These diagrams show expected sample accuracy as a function of confidence~\cite{reliability_diagram}. The blue bar shows the confidence of the bin and the orange bar shows the percentage correctness of prediction in that bin. If the model is perfectly calibrated, the bars align to form identity function. Any deviation from a perfect diagonal is a miscalibration.}
    \label{fig:calibration}
\end{figure*}

We observe that adding IOC constraints improve calibration error on the base NN architecture. We present the reliability diagrams~\cite{reliability_diagram} (presenting accuracy as a function of confidence) of three neural architectures and their convex counterparts in Figure~\ref{fig:calibration}. The sum of the difference between the blue bars and the orange bars represents the Expected Calibration Error. IOC constraints show improved calibration in all three architectures (with notable improvements in the case of MLP and AllConv). Better calibration further strengthens the case for IOC-NNs from the application perspective. 

\section{Conclusions}
We present a subclass of neural networks, where the output is a convex function of the input. We show that with minimal constraints, existing neural networks can be adopted to this subclass called Input Output Convex Neural Networks. With a set of carefully chosen experiments, we unveil that IOC-NNs show outstanding generalization ability and robustness to label noise while retaining adequate capacity. We show that in scenarios where the neural network gives a large generalization gap, IOC-NN can give better test performance. An alternate interpretation of our work can be self regularization (regularization through functional constraints). IOC-NN puts to rest the concerns around brute force memorization of deep neural networks and opens a promising horizon for the community to explore. We show that in the case of Multi-Layer-Perceptrons, IOC constraints improve accuracy, generalization, calibration, and robustness to noise, making an ideal proposition from a deployment perspective. The improved generalization, calibration, and robustness to noise are also observed in convolutional architectures while retaining the accuracy. In future work, we plan to investigate the use of IOC-NNs for recurrent architectures. Furthermore, we plan to explore the interpretability aspects of IOC-NNs and study the effect of convexity constraints on the generalization bounds.

\bibliographystyle{plain}
\bibliography{convexnn}

\begin{thebibliography}{10}

\bibitem{amos2017input}
Brandon Amos, Lei Xu, and J~Zico Kolter.
\newblock Input convex neural networks.
\newblock In {\em Proceedings of the 34th International Conference on Machine
  Learning-Volume 70}, pages 146--155. JMLR. org, 2017.

\bibitem{arpit2017closer}
Devansh Arpit, Stanislaw Jastrzebski, Nicolas Ballas, David Krueger, Emmanuel
  Bengio, Maxinder~S Kanwal, Tegan Maharaj, Asja Fischer, Aaron Courville,
  Yoshua Bengio, et~al.
\newblock A closer look at memorization in deep networks.
\newblock In {\em International Conference on Machine Learning}, pages
  233--242. PMLR, 2017.

\bibitem{bach2017breaking}
Francis Bach.
\newblock Breaking the curse of dimensionality with convex neural networks.
\newblock {\em The Journal of Machine Learning Research}, 18(1):629--681, 2017.

\bibitem{cifair_dataset}
Björn Barz and Joachim Denzler.
\newblock Do we train on test data? purging cifar of near-duplicates.
\newblock {\em Journal of Imaging}, 6:41, 06 2020.

\bibitem{boyd2004convex}
Stephen Boyd, Stephen~P Boyd, and Lieven Vandenberghe.
\newblock {\em Convex optimization}.
\newblock Cambridge university press, 2004.

\bibitem{burges1998a}
Chris~J.C. Burges.
\newblock A tutorial on support vector machines for pattern recognition.
\newblock {\em Data Mining and Knowledge Discovery}, 2:121--167, January 1998.

\bibitem{chen2018optimal}
Yize Chen, Yuanyuan Shi, and Baosen Zhang.
\newblock Optimal control via neural networks: A convex approach.
\newblock {\em arXiv preprint arXiv:1805.11835}, 2018.

\bibitem{chen2020input}
Yize Chen, Yuanyuan Shi, and Baosen Zhang.
\newblock Input convex neural networks for optimal voltage regulation.
\newblock {\em arXiv preprint arXiv:2002.08684}, 2020.

\bibitem{reliability_diagram}
Morris~H DeGroot and Stephen~E. Fienberg.
\newblock The comparison and evaluation of forecasters.
\newblock {\em The statistician}, pages 12--22, 1983.

\bibitem{friedman2000additive}
Jerome Friedman, Trevor Hastie, Robert Tibshirani, et~al.
\newblock Additive logistic regression: a statistical view of boosting (with
  discussion and a rejoinder by the authors).
\newblock {\em The annals of statistics}, 28(2):337--407, 2000.

\bibitem{calibration}
Chuan Guo, Geoff Pleiss, Yu~Sun, and Kilian Weinberger.
\newblock On calibration of modern neural networks.
\newblock {\em arXiv preprint arXiv:1706.04599}, 06 2017.

\bibitem{huang2017densely}
Gao Huang, Zhuang Liu, Laurens van~der Maaten, and Kilian~Q Weinberger.
\newblock Densely connected convolutional networks.
\newblock In {\em Proceedings of the IEEE Conference on Computer Vision and
  Pattern Recognition}, 2017.

\bibitem{10.1162/neco.1991.3.1.79}
Robert~A. Jacobs, Michael~I. Jordan, Steven~J. Nowlan, and Geoffrey~E. Hinton.
\newblock Adaptive mixtures of local experts.
\newblock {\em Neural Comput.}, 3(1):79–87, March 1991.

\bibitem{doi:10.1162/089976600300015367}
Wenxin Jiang.
\newblock The vc dimension for mixtures of binary classifiers.
\newblock {\em Neural Computation}, 12(6):1293--1301, 2000.

\bibitem{NIPS2014_5511}
Alex Kantchelian, Michael~C Tschantz, Ling Huang, Peter~L Bartlett, Anthony~D
  Joseph, and J.~D. Tygar.
\newblock Large-margin convex polytope machine.
\newblock In {\em Advances in Neural Information Processing Systems 27}, pages
  3248--3256. Curran Associates, Inc., 2014.

\bibitem{kent2016input}
Spencer Kent, Eric Mazumdar, Anusha Nagabandi, and Kate Rakelly.
\newblock Input-convex neural networks and posynomial optimization, 2016.

\bibitem{peicewise}
Anita Kripfganz and R.~Schulze.
\newblock Piecewise affine functions as a difference of two convex functions.
\newblock {\em Optimization}, 18(1):23--29, 1987.

\bibitem{krueger2017deep}
David Krueger, Nicolas Ballas, Stanislaw Jastrzebski, Devansh Arpit, Maxinder~S
  Kanwal, Tegan Maharaj, Emmanuel Bengio, Asja Fischer, and Aaron Courville.
\newblock Deep nets don't learn via memorization, 2017.

\bibitem{doi:10.1162/neco.2008.20.1.288}
Robert Legenstein and Wolfgang Maass.
\newblock On the classification capability of sign-constrained perceptrons.
\newblock {\em Neural Computation}, 20(1):288--309, 2008.

\bibitem{magnani2009convex}
Alessandro Magnani and Stephen~P Boyd.
\newblock Convex piecewise-linear fitting.
\newblock {\em Optimization and Engineering}, 10(1):1--17, 2009.

\bibitem{manwani2010learning}
Naresh Manwani and PS~Sastry.
\newblock Learning polyhedral classifiers using logistic function.
\newblock In {\em Proceedings of 2nd Asian Conference on Machine Learning},
  pages 17--30, 2010.

\bibitem{focal_loss}
Jishnu Mukhoti, Viveka Kulharia, Amartya Sanyal, Stuart Golodetz, Philip Torr,
  and Puneet Dokania.
\newblock Calibrating deep neural networks using focal loss.
\newblock {\em arXiv preprint arXiv:2002.09437}, 02 2020.

\bibitem{unifrom_conv}
Vaishnavh Nagarajan and J.~Kolter.
\newblock Uniform convergence may be unable to explain generalization in deep
  learning.
\newblock {\em arXiv preprint arXiv:1902.04742}, 02 2019.

\bibitem{neyshabur2017exploring}
Behnam Neyshabur, Srinadh Bhojanapalli, David McAllester, and Nati Srebro.
\newblock Exploring generalization in deep learning.
\newblock In {\em Advances in Neural Information Processing Systems}, pages
  5947--5956, 2017.

\bibitem{towards_overparameter}
Behnam Neyshabur, Zhiyuan Li, Srinadh Bhojanapalli, Yann LeCun, and Nathan
  Srebro.
\newblock Towards understanding the role of over-parametrization in
  generalization of neural networks.
\newblock {\em arXiv preprint arXiv: 1805.12076}, 05 2018.

\bibitem{neyshabur2015inductivebias}
Behnam Neyshabur, Ryota Tomioka, and Nathan Srebro.
\newblock In search of the real inductive bias: On the role of implicit
  regularization in deep learning.
\newblock {\em arXiv preprint arXiv:1412.6614}, 12 2014.

\bibitem{shah2019plume}
Kulin Shah, PS~Sastry, and Naresh Manwani.
\newblock Plume: Polyhedral learning using mixture of experts.
\newblock {\em arXiv preprint arXiv:1904.09948}, 2019.

\bibitem{sjoberg1995overtraining}
Jonas Sj{\"o}berg and Lennart Ljung.
\newblock Overtraining, regularization and searching for a minimum, with
  application to neural networks.
\newblock {\em International Journal of Control}, 62(6):1391--1407, 1995.

\bibitem{springenberg2014striving}
Jost~Tobias Springenberg, Alexey Dosovitskiy, Thomas Brox, and Martin
  Riedmiller.
\newblock Striving for simplicity: The all convolutional net.
\newblock {\em arXiv preprint arXiv:1412.6806}, 2014.

\bibitem{Takacs2007}
Gábor Takács and Béla Pataki.
\newblock Lower bounds on the vapnik-chervonenkis dimension of convex polytope
  classifiers.
\newblock In {\em 2007 11th International Conference on Intelligent Engineering
  Systems}, pages 145 -- 148, 06 2007.

\bibitem{zhang2016understanding}
Chiyuan Zhang, Samy Bengio, Moritz Hardt, Benjamin Recht, and Oriol Vinyals.
\newblock Understanding deep learning requires rethinking generalization.
\newblock {\em arXiv preprint arXiv:1611.03530}, 2016.

\end{thebibliography}

\title{Appendix}
\maketitle
\vspace{-5mm}
This document contains the supplementary material to support the main text. The contents of this document are:
\begin{itemize}
    \item Appendix A: Using non-convex activations in the last layer of Input Output Convex Neural Networks (IOC-NNs)
    \item Appendix B: Optimization algorithm for training IOC-NNs
    \item Appendix C: Confidence intervals of classification accuracies for neural networks and IOC-NNs
    \item Appendix D: Discussion on capacity of IOC-NNs
    \item Appendix E: Partition of input space by MoEs of IOC-NNs 
    
\end{itemize}
\appendix
\section{Using Non-Convex Activations to Facilitate Training}
In a multi-class classification setting, the softmax function is widely used to get a joint probability distribution over the output classes. This facilitates training with categorical cross-entropy loss. The softmax layer distorts input output convexity, but the decision boundary remains unchanged even after applying softmax in the last layer. Each of the pre-softmax output of IOC-NNs is convex with respect to the inputs. The classification decision is determined by the order (rank) of these values, as this rank remains unaffected with the application of the softmax function. At inference time, we compute $\argmax$ of convex functions (pre-softmax layer).
\begin{equation} \footnotesize
    \argmax(Y) = \argmax(softmax(Y))
\end{equation}
A similar approach is taken for modelling independent probabilities of labels in binary classification setting. Sigmoid activation is widely used to model the probability $p(y=y_n|x)$. For IOC-NN, at inference, the predicted value is $ y = \sigma(f(x))$, where $f(x)$ is convex. Hence, decision boundary is:

\begin{equation} \footnotesize
    \sigma(f(x)) = 0.5 ,
\end{equation}    
which is equivalent to 
\begin{equation} \footnotesize
f(x) = \sigma^{-1}(0.5)
\end{equation}
where, $f(x)$ is convex. Since $\sigma^{-1}(0.5)$ is a constant, the decision boundary is convex.

\section{Optimization Algorithm for Training IOC-NNs}
The only architectural constraint in designing an IOC-NN is the choice of a convex and non-decreasing activation function. Furthermore, any feed-forward neural architecture can be trained as IOC-NN by adding two steps to the optimization algorithm. For example, the constrained version of the vanilla stochastic gradient descent algorithm is shown in Algorithm ~\ref{alg:the_alg}.

\setlength{\algomargin}{2.0pt}

\begin{algorithm}[]
\label{alg:the_alg}
\SetAlgoLined
\vspace{1mm}
\KwIn{Training data $S$, Labels $L$, learning rate $\eta$, constant $\epsilon$, $\theta_0$ = ($w_0{^n_{ij}}$, $b_0{^n_{i}}$)}
\KwOut{$\Theta = (w^*,b^*)$}
 initialize: $\theta = \theta_0$\;
 
 \While{stopping criteria not met}{
  \For{i in 1:len(S) / batch\_size }{
  sample$ (x_{batch},y_{batch}) \in (S,L)$\;
  $L \leftarrow Div (f_{\theta}(x_{batch}),y_{batch})$\; 
  $w \leftarrow w - \eta(\frac{\partial}{\partial w} L)$\;
  $b \leftarrow b - \eta(\frac{\partial}{\partial b} L)$\; 
  \For {layer $\in {2:k}$}{
  \If{$w < 0$ }{
  $w \leftarrow e^{w-\epsilon}$;
  }{
  }}}
  check stopping criteria
  }

\caption{Algorithm to train a k layer IOC-NN}
\end{algorithm}

$\epsilon$ is used to bring down the value of the updated weights post exponentiation for negative values close to zero. Values near zero, post exponentiation, will be close to one, and $\epsilon$ helps keep them close to zero. We use $\epsilon = 5$ across all experiments. To summarize, the only two additional steps required to train an IOC-NN are condition (to check the sign of updated weights) and exponentiation. We can develop IOC constrained versions for all optimization algorithms that are used in training neural networks. 

\section{Confidence Intervals of Classification Accuracies for Neural Networks and IOC-NNs}

We run MLP, AllConv and Densenet on CIFAR-10 data five times.  We report the accuracy in prediction and the confidence intervals obtained over five runs in Table \ref{table:intervals}. We report the train and test accuracies along with the generalization gap. The table confirms the reliability of the results in the main document. The generalization gap of IOC-NNs are consistently lower than neural networks. Also, compared to base MLP the convex counterpart gives higher test accuracy.
\begin{table}[t]
\centering
\caption{.}
\vspace{2mm}
\label{table:intervals}
\small\addtolength{\tabcolsep}{+1.0pt}
\begin{tabular}{|c|c|c|c|c|c|c|}
\hline
& \multicolumn{3}{c|}{NN} & \multicolumn{3}{c|}{IOC-NN} \\ \cline{2-7} 
\multicolumn{1}{|c|}{}   & train      & test    & Gen. error      & train        & test   &Gen. error        \\ \hline
MLP      & 97.99 $\pm$ 0.18 & 63.83 $\pm$ 0.12  & 34.1 & 73.27 $\pm$ 0.41 & 69.89 $\pm$ 0.21 & 3.38 \\ \cline{2-7}   \hline
All Conv    & 99.12 $\pm$ 1.03   & 92.21 $\pm$ 0.18   & 7.09  & 92.61 $\pm$ 1.45 & 90.34 $\pm$ 0.20 & 2.27 \\ \cline{2-7} 
\hline
DenseNet & 99.08 $\pm$ 0.75 & 93.86 $\pm$ 0.46     & 5.22         & 93.8 $\pm$ 0.83     & 90.43 $\pm$ 0.59 & 3.37     \\ \cline{2-7} 
\hline
\end{tabular}
\end{table}

\section{Discussion on Capacity of IOC-NNs}

The decision boundaries of real-world classification problems are often not convex. Each of the single outputs of IOC-NN is convex with respect to the inputs by design. However, it can still learn arbitrarily complex decision boundaries. Kripfganz~\etal \cite{peicewise} show that we can represent any piecewise linear function as a difference of two piecewise linear convex functions. Using ReLU as an activation limits the function classes that convex networks can learn. For instance, they cannot learn identity mapping \cite{amos2017input}. A simple architectural change of using ELU overcomes this issue and hence increases the capacity of IOC-NNs. Results in the main text show empirical evidence of this increase in capacity. Following this direction in the future, we would like to explore the proof for universal approximation for IOC-NNs. 

\begin{figure}[!h]
    \centering
    \begin{tabular}[t]{ccccc}
    \hspace{-1.5em}
        \includegraphics[width=0.25\linewidth]{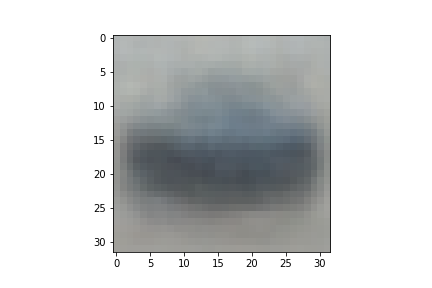}& \hspace{-2.5em} \includegraphics[width=0.25\linewidth]{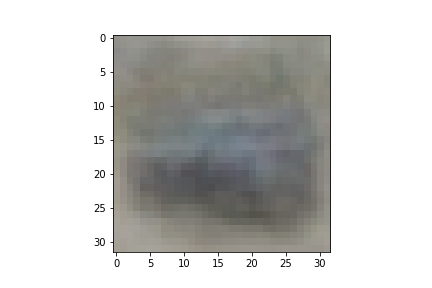}& \hspace{-2.5em} \includegraphics[width=0.25\linewidth]{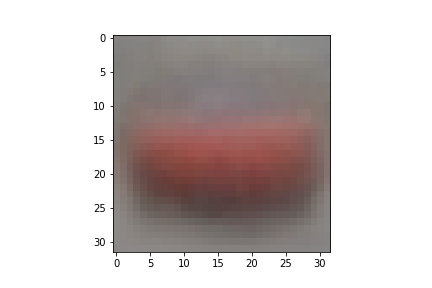}& \hspace{-2.5em} \includegraphics[width=0.25\linewidth]{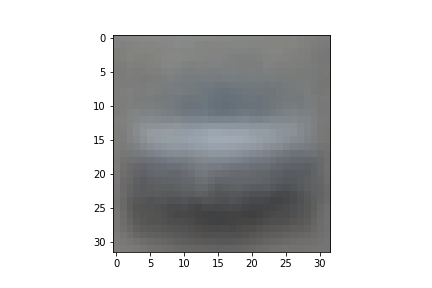}& \hspace{-2.5em}
        \includegraphics[width=0.25\linewidth]{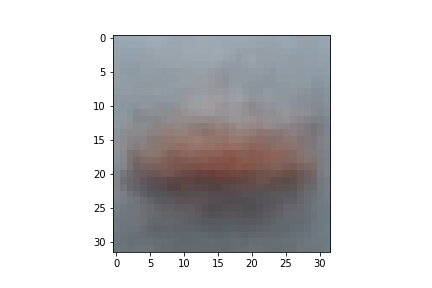}
        \\
        \hspace{-1.5em}
        \includegraphics[width=0.25\linewidth]{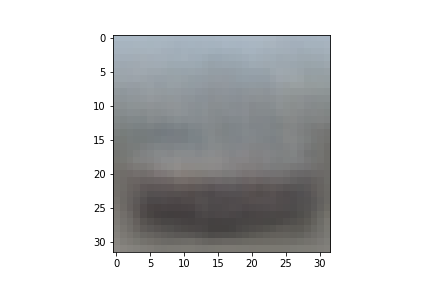}& \hspace{-2.5em} 
        \includegraphics[width=0.25\linewidth]{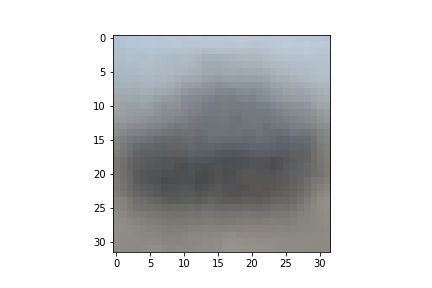}& \hspace{-2.5em} \includegraphics[width=0.25\linewidth]{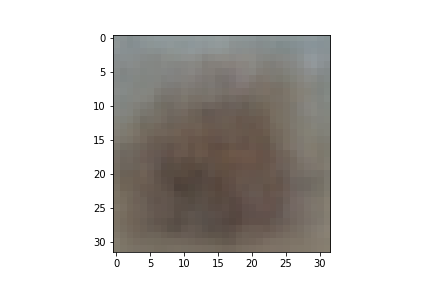}& \hspace{-2.5em}
        \includegraphics[width=0.25\linewidth]{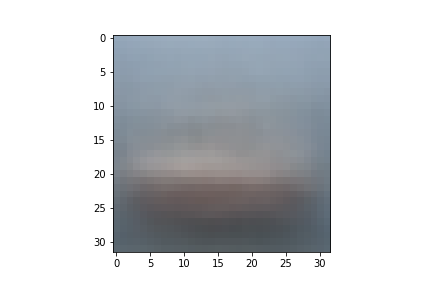}& \hspace{-2.5em} \includegraphics[width=0.25\linewidth]{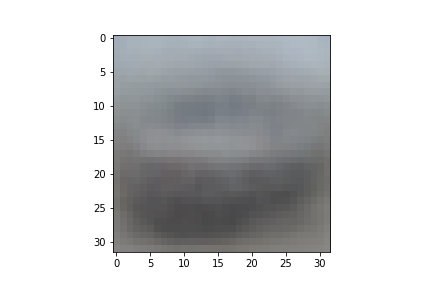} 
        
    \end{tabular}
    \vspace{1mm}
    \caption{The figure shows the mean image of ten clusters within the class 'Automobile'. Each cluster corresponds to the expert picked by the gate. These partitions signify the difference in the data points on which each of the binary convex models gained expertise. }
    \label{fig:partition}
\end{figure}

\section{Partition of Input space by MoEs of IOC-NNs}
Real-world data often lies in a sparse, high-dimensional space and neural networks fit complex boundaries over them. These complex boundaries make the classification results challenging to interpret. In section 3.2 of the main text, we demonstrate on toy data, how an ensemble of binary IOC-NNs creates meaningful partitions of input space. Gating network of MoEs of IOC-NNs partition input space into mutually exclusive and exhaustive subspaces. The partition chosen by the gate for each expert is a portion where a convex boundary is enough to make a reasonable prediction. We train ten, three layered single output IOC-MLPs using gated EM strategy on the two-class setting of CIFAR-10 as explained in section 4: 'Training ensembles of binary experts'.     
Figure ~\ref{fig:partition} shows the mean image of ten partitions of class 'Automobile'. Each partition corresponds to one of the ten experts chosen by the gate. Each of the IOC-NN has expertise on clusters with aspects that can lead to some real-world interpretation. For instance, an expert is chosen by the gate to classify images showing the front view of red cars from similar images, while another expert is picked for side view images. Similar clusters can be seen corresponding to different colors and orientations. In the future, we would like to explore this property of MoE of binary IOC-NNs towards improving the interpretability of decisions made by neural networks.   
\end{document}